\documentclass[DTMColor]{ROB-New}

\usepackage[utf8]{inputenc} 
\usepackage[T1]{fontenc}    
\usepackage{hyperref}       
\usepackage{url}            
\usepackage{booktabs}       
\usepackage{amsfonts}       
\usepackage{nicefrac}       
\usepackage{microtype}      
\usepackage{amssymb,amsfonts,amsmath,amscd}
\usepackage[printonlyused,withpage]{acronym}
\usepackage{bm}

\begin{document}

\authormark{Barfoot, Forbes, D'Eleuterio}

\articletype{}

\jnlPage{1}{00}
\jyear{2021}
\jdoi{10.1017/xxxxx}

\title{Vectorial Parameterizations of Pose}

\author[1]{Timothy D. Barfoot\hyperlink{corr}{*}}
\address[1]{University of Toronto Institute for Aerospace Studies}

\author[2]{James R. Forbes}
\address[2]{Department of Mechanical Engineering, McGill University}

\author[1]{Gabriele M. T. D'Eleuterio}

\address{\hypertarget{corr}{*}Corresponding author. \email{tim.barfoot@utoronto.ca}}


\graphicspath{{figs/}}
\newcommand{\diag}{\textup{diag}}
\newcommand{\bcf}{\;\mbox{\boldmath ${\cal F}$\unboldmath}}
\newcommand{\bbm}{\begin{bmatrix}}
\newcommand{\ebm}{\end{bmatrix}}
\newcommand{\mbf}{\mathbf}
%
%
\newcommand{\mbs}[1]{{\boldsymbol{#1}}}
\def\ep{\epsilon}
\def\la{\lambda}
\def\om{\omega}
\newcommand{\qc}[1]{{#1}^+}
\newcommand{\qo}[1]{{#1}^\oplus}
\newcommand{\qi}[1]{{#1}^{\scalebox{0.5}{$-1$}}}
%
%
%
\renewcommand{\Vec}[1]{\underrightarrow{#1}}
\newcommand{\beq}{\begin{equation}}
\newcommand{\eeq}{\end{equation}}
\newcommand{\bdis}{\begin{displaymath}}
\newcommand{\edis}{\end{displaymath}}
\newcommand{\bomg}{\mbox{\boldmath $\omega$\unboldmath}}
\newcommand{\bfr}{\mbf{r}}
\newcommand{\bfa}{\mbf{a}}
\newcommand{\bfone}{\mbf{1}}
\newcommand{\bfs}{\mbf{s}}
\newcommand{\bfzero}{\mbf{0}}
\newcommand{\bfC}{\mbf{C}}
\newcommand{\bfS}{\mbf{S}}
\newcommand{\bu}{{\scalebox{0.5}{\mbox{$\bullet$}}}}
\newcommand{\ci}{{\scalebox{0.5}{\mbox{$\circ$}}}}
\newcommand{\cjda}[1]{{#1}^\bu}
\newcommand{\cjdb}[1]{{#1}^\ci}
\newcommand{\cjdaa}[1]{{#1}^{\bu\bu}}
\newcommand{\cjdbb}[1]{{#1}^{\ci\ci}}
\def \bfth{\boldsymbol{\theta}}
\newcommand{\beqn}[1]{\begin{subequations}\label{eq:#1}\begin{eqnarray}}
\newcommand{\eeqn}{\end{eqnarray}\end{subequations}}
\newcommand{\est}[1]{\hat{#1}}
\newcommand{\pri}[1]{\check{#1}}
%
\newcommand{\wdg}{\wedge}
\newcommand{\uwdg}{\vee}
\newcommand{\Wdg}{\curlywedge}
\newcommand{\uWdg}{\curlyvee}
%
\newcommand{\pfs}{\odot}
\newcommand{\psf}{\circledcirc}
\newcommand{\Tbig}{\;\mbox{\boldmath ${\cal T}$\unboldmath}}
\newcommand{\nTbig}{\widetilde{\mbs{\mathcal{T}}}}
\newcommand{\Tsmall}{\mbf{T}}
\newcommand{\nTsmall}{\widetilde{\mbf{T}}}
\newcommand{\Rbig}{\mbs{\mathcal{R}}}
\newcommand{\Jsmall}{\mbf{J}}
\newcommand{\Jbig}{\mbs{\mathcal{J}}}
\newcommand{\lop}{\langle \! \langle}
\newcommand{\rop}{\rangle \! \rangle}

\newcommand{\mbc}[1]{{\mathcal{#1}}}
\newcommand{\ip}[2]{{\left< #1,#2 \right>}}
\newcommand{\op}[2]{{\left. #1 \, \right> \! \left< \,#2 \right.}}
\renewcommand{\vec}{\mbox{vec}}
\newcommand{\vech}{\mbox{vech}}
\newcommand{\mat}{\mbox{mat}}
\newcommand{\matf}{\mbox{matf}}
\newcommand{\sym}{\mbox{sym}}
\newcommand{\tr}{\mbox{tr}}
\newcommand{\rank}{\mbox{rank}}
\newcommand{\norm}[1]{\,\downarrow\!{#1}}
\newcommand{\length}[1]{\left\|{#1}\right\|}

\renewcommand{\labelitemi}{--}
\renewcommand{\labelitemii}{--}

\newcommand{\change}[1]{\color{red}#1\color{black}}


\abstract{Robotics and computer vision problems commonly require handling rigid-body motions comprising translation and rotation -- together referred to as {\em pose}.  In some situations, a {\em vectorial} parameterization of pose can be useful, where elements of a vector space are surjectively mapped to a matrix Lie group.  For example, these vectorial representations can be employed for optimization as well as uncertainty representation on groups.  The most common mapping is the matrix exponential, which maps elements of a Lie algebra onto the associated Lie group.  However, this choice is not unique.  It has been previously shown how to characterize all such vectorial parameterizations for $SO(3)$, the group of rotations.  Some results are also known for the group of poses, where it is possible to build a family of vectorial mappings that includes the matrix exponential as well as the Cayley transformation.  We extend what is known for these pose mappings to the $4 \times 4$ representation common in robotics, and also demonstrate three different examples of the proposed pose mappings:  (i) pose interpolation, (ii) pose servoing control, and (iii) pose estimation in a pointcloud alignment problem.  In the pointcloud alignment problem our results lead to a new algorithm based on the Cayley transformation, which we call {\em CayPer}.}

\maketitle


\section{Introduction}

Describing rigid-body motion is a fundamental concept in robotics, computer vision, computer graphics, augmented reality, classical mechanics, and beyond.  From \citet{mozzi1763} and \citet{chasles1830} we have the {\em Mozzi-Chasles theorem}, which says that the most general rigid-body displacement can be produced by a translation along a line (the {\em screw axis}) followed by (or preceeded by) a rotation about that line.  A complete screw theory was later developed by \citet{ball1900}.  \citet{murray94} popularized the modern approach to describing rigid-body motion (in the context of manipulator robotics) using the {\em special Euclidean group}, where screw motions are elegantly described by mapping elements of the Lie algebra onto the Lie group through the matrix exponential.  It is through this sequence of historical events that multiple communities have come to think of the matrix exponential as the {\em canonical} way to map a six-degree-of-freedom {\em vector} to a rigid-body {\em pose}.  Optimization and uncertainty representation for pose quantities now also commonly make use of matrix Lie group tools and the exponential map \citep{chirikjian09,barfoot_tro14,barfoot_ser17}.

Yet, there are many representations of three-dimensional {\em rotations} in active use today:  Euler angles, unit-length quaternions (a.k.a., Euler parameters), rotation vectors, Cayley-Gibbs-Rodrigues parameters, modified Rodrigues parameters, and many more \citep{hughes86}.  Of these, only the {\em rotation vector} (i.e., angle multiplied with a unit-length axis) maps to an element of the {\em special orthogonal group} (i.e., a rotation matrix) through the matrix exponential.
However, \citet{bauchau03} synthesized the class of {\em vectorial} parameterizations of rotation.  In particular, they elegantly describe a general mapping from a three-degree-of-freedom vector to a rotation matrix that encompasses many commonly employed representations of rotation, including Cayley-Gibbs-Rodrigues and modified Rodrigues parameters.

\citet{bauchau03b} and later \citet{bauchau11b} expanded the idea of a general vector mapping to the special Euclidean group representing poses, where they focus on the $6 \times 6$ {\em adjoint} representation of pose, $\mbox{Ad}(SE(3))$.  \citet{bauchau11} provides an excellent summary of both the rotation and adjoint pose vectorial mappings.  Interestingly, these works do not present the general vector mappings for the common $4 \times 4$ representation of pose, $SE(3)$, although some works do explore special cases \citep{borri00,selig07}.  We revisit the idea of a {\em vectorial parameterization of pose}, extending what is known in a few specific ways.  First, we explore the more common $4 \times 4$ representation, $SE(3)$, showing that for every parameterization of rotation, there are infinite possibilities for the parameterization of pose having different properties from which one might choose; each has a different coupling between the translational and rotational variables.  Second, we show that our general vector mapping provides the special cases of the exponential map and the {\em Cayley transformation} \citep{cayley1846} for $SE(3)$, which has been previously discussed in the literature \citep{borri00, selig07}.  Finally, we hope that this paper serves to further shine a light on the important long-term efforts of \citet{bauchau11} on the vectorial parameterizations of rotations and poses, which we believe is of great interest to the robotics community.

Further to our extensions of the vectorial parameterization of pose, which is a theoretical contribution, we also demonstrate how to use the vectorial parameterization of pose in various canonical robotics applications. In particular, the classic problems of pose interpolation, pose control, and pose estimation in a pointcloud alignment framework, are each addressed using different vectorial parameterizations of pose. We show using an alternative pose parameterization over the classic matrix exponential can lead to different results in terms of performance and robustness. 

The paper is organized as follows.  Section~\ref{sec:math} provides some brief mathematical background.  Section~\ref{sec:vecmap} introduces our vectorial parameterization of pose and discusses several properties thereof.  While our main contribution is the theory, Section~\ref{sec:apps} provides three applications highlighting that different pose mappings may be beneficial in different situations.  Section~\ref{sec:confu} concludes the paper.

\section{Mathematical Preliminaries}\label{sec:math}

We briefly review some key concepts and notation \citep{barfoot_ser17} to prepare for what follows.

\subsection{Matrix Lie Groups}

The {\em special orthogonal group}, representing rotations, is the set of valid rotation matrices:
\begin{equation}
\label{eq:SO3}
SO(3) = \left\{  \mbf{C} \in \mathbb{R}^{3\times3} \; | \; \mbf{C} \mbf{C}^T = \mbf{1}, \mbox{det} \,\mbf{C} = 1 \right\},
\end{equation}
where $\mbf{1}$ is the identity matrix.  
It is common to map a vector, $\mbs{\phi} \in \mathbb{R}^3$, to a rotation matrix, $\mbf{C}$, through the matrix exponential,
\begin{equation}
\mbf{C}(\mbs{\phi}) = \exp\left( \mbs{\phi}^\wdg \right),
\end{equation}
where $(\cdot)^\wdg$ is the skew-symmetric operator,
\begin{equation}
\mbs{\phi}^\wdg = \bbm \phi_1 \\ \phi_2 \\ \phi_3 \ebm^\wdg = \bbm 0 & -\phi_3 & \phi_2 \\ \phi_3 & 0 & -\phi_1 \\ -\phi_2 & \phi_1 & 0 \ebm.
\end{equation}
The mapping is surjective-only, meaning every $\mbf{C}$ can be produced by many different values for $\mbs{\phi}$.

The {\em special Euclidean group}, representing \index{poses} poses (i.e., translation and rotation), is the set of valid  transformation matrices:
\begin{equation}
\label{eq:se3}
SE(3) = \left\{  \mbf{T} = \bbm \mbf{C} & \mbf{r} \\ \;\,\mbf{0}^T & 1 \ebm \in \mathbb{R}^{4\times4} \; \Biggl| \; \mbf{C} \in SO(3), \, \mbf{r} \in \mathbb{R}^3 \right\}.
\end{equation}
It is again common to map a vector, $\mbs{\xi} \in \mathbb{R}^6$, to a transformation matrix, $\mbf{T}$, through the matrix exponential,
\begin{equation}
\mbf{T}(\mbs{\xi}) = \exp\left( \mbs{\xi}^\wdg \right),
\end{equation}
where
\begin{equation}
\mbs{\xi}^\wdg = \bbm \mbs{\rho} \\ \mbs{\phi} \ebm^\wdg = \bbm \mbs{\phi}^\wdg & \mbs{\rho} \\ \mbf{0}^T & 0 \ebm.
\end{equation}
As is common practice \citep{barfoot_ser17}, we have broken the pose vector, $\mbs{\xi}$, into a translational component, $\mbs{\rho}$, and a rotational component, $\mbs{\phi}$.
The mapping is also surjective-only, meaning every $\mbf{T}$ can be produced by many different values for $\mbs{\xi}$.

Finally, the {\em adjoint} of pose is given by
\begin{equation}\label{eq:se3adjointmap1}
\Tbig(\mbs{\xi}) = \mbox{Ad}\left( \mbf{T} \right) = \bbm \mbf{C}(\mbs{\phi}) & \mbf{r}^\wdg \mbf{C}(\mbs{\phi}) \\ \mbf{0} & \mbf{C}(\mbs{\phi}) \ebm,
\end{equation}
which is now $6 \times 6$.  We will refer to the set of adjoints as $\mbox{Ad}(SE(3))$.  We can map a vector, $\mbs{\xi} \in \mathbb{R}^6$, to an adjoint transformation matrix again through the matrix exponential map:
\begin{equation}
\Tbig(\mbs{\xi}) = \exp\left( \mbs{\xi}^\Wdg \right),
\end{equation}
where
\begin{equation}
\mbs{\xi}^\Wdg = \bbm \mbs{\rho} \\ \mbs{\phi} \ebm^\Wdg = \bbm \mbs{\phi}^\wdg & \mbs{\rho}^\wdg \\ \mbf{0} & \mbs{\phi}^\wdg \ebm.
\end{equation}
The mapping is again surjective-only, meaning every $\Tbig$ can be produced by many different values for $\mbs{\xi}$.

\subsection{Key Identities}

We will be exploring mappings (including the exponential map) from vectors to Lie groups and will have occasion to work with series expressions of rotation and transformation matrices.  Each group has associated with it an identity that can be used to limit the number of terms in such series, related to the Cayley-Hamilton theorem.

For rotations, we have that the characteristic equation of $\mbs{\phi}^\wdg$ is 
\begin{equation}
\left|  \lambda \mbf{1} - \mbs{\phi}^\wdg \right| = \lambda ( \lambda^2 + \underbrace{\phi_1^2 + \phi_2^2 + \phi_3^2}_{\phi^2} ) = \lambda^3 + \phi^2 \lambda = 0,
\end{equation}
where $\lambda$ are the eigenvalues of $\mbs{\phi}^\wdg$.
From the Cayley-Hamilton theorem we can claim that
\begin{equation} \label{eq:ch1}
\mbs{\phi}^{\wdg^3} + \phi^2 \mbs{\phi}^\wdg \equiv \mbf{0},
\end{equation}
since $\mbs{\phi}^\wdg$ must satisfy its own characteristic equation.

For poses, we have that the characteristic equation of $\mbs{\xi}^\wdg$ is
\begin{equation}\label{eq:se3char}
\left|  \lambda \mbf{1} - \mbs{\xi}^\wdg \right| = \left| \begin{matrix} \lambda \mbf{1} - \mbs{\phi}^\wdg & -\mbs{\rho} \\ \mbf{0}^T & \lambda \end{matrix} \right| = \lambda^4 + \phi^2 \lambda^2 = 0.
\end{equation}
Again, from the Cayley-Hamilton theorem we can claim that
\begin{equation}\label{eq:ch2}
\mbs{\xi}^{\wdg^4} + \phi^2 \mbs{\xi}^{\wdg^2} \equiv \mbf{0},
\end{equation}
since $\mbs{\xi}^\wdg$ must satisfy its own characteristic equation.

Finally, for adjoint poses we have that the characteristic equation of $\mbs{\xi}^\Wdg$ is
\begin{equation}\label{eq:adse3char}
\left|  \lambda \mbf{1} - \mbs{\xi}^\Wdg \right| = \left| \begin{matrix} \lambda \mbf{1} - \mbs{\phi}^\wdg & -\mbs{\rho}^\wdg \\ \mbf{0} & \lambda \mbf{1} - \mbs{\phi}^\wdg \end{matrix} \right| =  \left(  \lambda^3 + \phi^2 \lambda  \right)^2 =  \lambda^6 + 2 \phi^2 \lambda^4 + \phi^4 \lambda^2 = 0.
\end{equation}
We could again employ the Cayley-Hamilton theorem to create a similar identity to the other two cases, but it turns out that the {\em minimal polynomial} of $\mbs{\xi}^\Wdg$ is actually \citep{barfoot_ser17, deleuterio_tro21},
\begin{equation}\label{eq:adid}
\mbs{\xi}^{\Wdg^5} + 2\phi^2 \mbs{\xi}^{\Wdg^3} + \phi^4 \mbs{\xi}^\Wdg \equiv \mbf{0},
\end{equation}
which is one order lower than the characteristic equation in this case.  This is because both the algebraic and geometric multiplicities of eigenvalue $\lambda=0$ in~\eqref{eq:adse3char} are two (two Jordan blocks of size one), allowing us to drop one copy when constructing the minimal polynomial; in~\eqref{eq:se3char} the algebraic multiplicity of $\lambda = 0$ is two, while the geometric multiplicity is one (one Jordan block of size two).

\section{Vector Mappings}\label{sec:vecmap}

In this section, we will show that there exist several options for mapping vectors onto the two most common matrix Lie groups used in robotics and computer vision.  Our approach builds on the work of \citet{bauchau03}, who worked this out for $SO(3)$, the group of rotations and \citet{bauchau03b,bauchau11} who extended this work to $\mbox{Ad}(SE(3))$, the adjoint representation of pose.  We show how to extend the rotation work to $SE(3)$ directly.

\subsection{Vector Mappings for Rotations}

It is well known that there exist many different ways to parameterize the group of three-dimensional rotations, $SO(3)$.  Naturally, such rotations have three degrees of freedom.  Representations with three parameters are known to always have singularities while representations with more than three parameters have constraints to keep the degrees of freedom at three \citep{stuelpnagel1964, hughes86}.  \citet{bauchau03} synthesized previous work on three-parameter representations that can be considered {\em vectors};  they show that all such parameterizations can be written as
\begin{equation}\label{eq:so3def}
\mbs{\phi} = \phi(\varphi) \, \mbf{a} \in \mathbb{R}^3,
\end{equation}
where $\mbf{a}$ is the unit-length axis of rotation, $\varphi$ the angle of rotation, and $\phi(\varphi)$ a {\em generating function}.  The scalar generating function is an odd function of the rotation angle that satisfies
\begin{equation}\label{eq:genprop}
\lim_{\varphi \rightarrow 0} \frac{\phi(\varphi)}{\varphi} = \kappa, 
\end{equation}
with $\kappa$ a real normalization constant.  We will assume $\kappa = 1$ in what follows as this makes it easier to compare different generating functions directly.  The generating function is essentially a nonlinear warping of one angle, $\varphi$, to new angle, $\phi$.

\begin{figure*}[t]
\centering
\includegraphics[width=\textwidth]{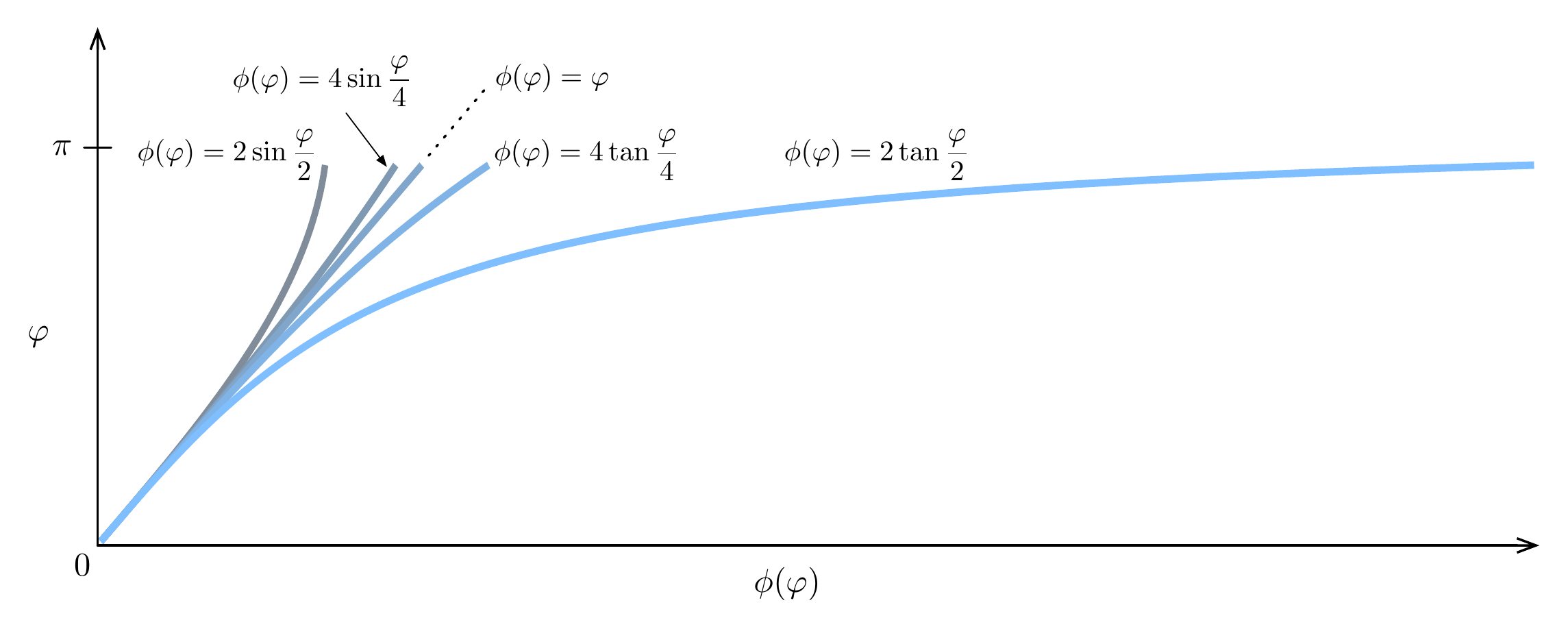}
\caption{Example generating functions, $\phi(\varphi)$, that map rotation angle, $\varphi$ to different vector representations of rotation and pose.  The independent variable is shown on the vertical axis to save space.  We see that the $2\tan\frac{\varphi}{2}$ function is quite a bit more extreme than the others}
\label{fig:genfunc}
\end{figure*}

Some common examples of vector parameterizations of rotation are,
\begin{equation}
\underbrace{\mbs{\phi} = \varphi \mbf{a}}_{\mbox{rotation vector}}, \quad \underbrace{\mbs{\phi} = 2 \tan \frac{\varphi}{2} \mbf{a}}_{\mbox{Cayley-Gibbs-Rodrigues}},  \quad \underbrace{\mbs{\phi} = 4 \tan \frac{\varphi}{4} \mbf{a}}_{\mbox{modified Rodrigues}},
\end{equation}
where we have ensured $\kappa = 1$ in~\eqref{eq:genprop} by including normalization constants where necessary.
Notably, the Euler angle sequences, a common three-parameter representation, do not fit the vector definition since they are not parallel to the axis of rotaton.  Vector parameterizations come along with singularities but these will not be a roadblock for the way in which we intend to use them.  Figure~\ref{fig:genfunc} depicts several generating functions.

\citet{bauchau03} show that the corresponding rotation matrix, $\mbf{C} \in SO(3)$, can be constructed from the general vector parameterization as follows:
\begin{equation}\label{eq:so3map}
\mbf{C}(\mbs{\phi}) = \mbf{1} + \frac{\nu^2}{\ep}\mbs{\phi}^\wdg + \frac{\nu^2}{2}\mbs{\phi}^{\wdg^2},
\end{equation}
where
\begin{equation}
\nu(\varphi) = \frac{2 \sin(\varphi/2)}{\phi(\varphi)}, \quad \ep(\varphi) = \frac{2 \tan(\varphi/2)}{\phi(\varphi)}.
\end{equation}
The group-mapped-from-vector function is a surjective-only mapping from $\mathbb{R}^3$ to $SO(3)$, which means every rotation matrix can be generated by many vectors (owing to the rotation angle wrapping around).  

The expression for the $SO(3)$ Jacobian, $\mbf{J}(\mbs{\phi})$, that maps the derivative of the vector parameters to angular velocity, 
\begin{equation}
\mbs{\omega} = \mbf{J}(\mbs{\phi}) \dot{\mbs{\phi}},
\end{equation}
is given by \citep{bauchau03}
\begin{equation}\label{eq:so3jac}
\mbf{J}(\mbs{\phi}) = \mu \mbf{1} + \frac{\nu^2}{2} \mbs{\phi}^\wdg + \frac{1}{\phi^2}\left( \mu - \frac{\nu^2}{\ep}\right) \mbs{\phi}^{\wdg^2},
\end{equation}
where
\begin{equation}
\mu(\varphi) = \left(\frac{d\phi(\varphi)}{d\varphi}\right)^{-1}.
\end{equation}
The inverse of this Jacobian is given by \citep{bauchau03}
\begin{equation}\label{eq:so3jacinv}
\mbf{J}(\mbs{\phi})^{-1} = \frac{1}{\mu} \mbf{1} - \frac{1}{2} \mbs{\phi}^\wdg - \frac{1}{\phi^2} \left( \frac{1}{\ep} - \frac{1}{\mu} \right)\mbs{\phi}^{\wdg^2}, 
\end{equation}
which we will require when dealing with $SE(3)$.

As the rotation angle, $\varphi$, becomes small, we have the following infinitesimal expressions:
\begin{equation}
\mbf{C}(\mbs{\phi}) \approx \mbf{1} + \mbs{\phi}^\wdg, \quad \mbf{J}(\mbs{\phi}) \approx \mbf{1} + \frac{1}{2} \mbs{\phi}^\wdg.
\end{equation}
These will become important when considering perturbations.

The result from this section is general, but it will be instructive to consider some specific examples.  Table~\ref{tab:so3vecparams} provides some common vector parameterizations and their associated mappings to $SO(3)$.  The first three mappings are appealing in that they do not require calculating the angle of rotation, $\varphi$, and instead can be built directly from $\mbs{\phi}$.

\begin{table*}[t]
\centering
\caption{Example vector parameterizations of rotation and the associated mappings to $SO(3)$.  Note the connection to the Cayley transformation in the $\tan$ parameterizations.  The entries in the table are ordered from steepest to shallowest generating function (right to left in Figure~\ref{fig:genfunc})}
\begin{tabular}{lcccc}
\hline
 & $\phi(\varphi)$ & $\mbf{C}(\mbs{\phi})$  & validity range \\ \hline 
Cayley-Gibbs-Rodrigues & $2 \tan \frac{\varphi}{2}$ & $\left( \mbf{1} - \frac{1}{2}\mbs{\phi}^\wdg \right)^{-1} \left( \mbf{1} + \frac{1}{2}\mbs{\phi}^\wdg \right)$ &  $| \varphi |< \pi$ \\
  modified Rodrigues & $4 \tan \frac{\varphi}{4}$ &  $\left( \mbf{1} - \frac{1}{4}\mbs{\phi}^\wdg \right)^{-2} \left( \mbf{1} + \frac{1}{4}\mbs{\phi}^\wdg \right)^2$ & $| \varphi | < 2\pi$ \\
rotation vector & $\varphi $ & $\exp\left(\mbs{\phi}^\wdg \right)$ & $| \varphi | < 2\pi$ \\
Bauchau-Trainelli & $4\sin\frac{\varphi}{4}$ &  $\mbf{1} + \cos\frac{\varphi}{2} \cos\frac{\varphi}{4}\mbs{\phi}^\wdg + \frac{1}{2} \cos^2 \frac{\varphi}{4}\mbs{\phi}^{\wdg^2}$ & $| \varphi | < 2\pi$ \\
 Euler-Rodrigues & $2 \sin\frac{\varphi}{2}$ & $\mbf{1} + \cos\frac{\varphi}{2} \mbs{\phi}^\wdg + \frac{1}{2} \mbs{\phi}^{\wdg^2}$ & $| \varphi | < \pi$  \\ \hline
\end{tabular}
\label{tab:so3vecparams}
\end{table*}

\subsection{Vector Mappings for Poses}

We next extend the idea of generalized vector mappings to $SE(3)$.  As pose changes have six degrees of freedom, our vector parameterization will take the form
\begin{equation}
\mbs{\xi} = \bbm \mbs{\rho} \\ \mbs{\phi} \ebm \in \mathbb{R}^6,
\end{equation}
where the rotational parameters, $\mbs{\phi} \in \mathbb{R}^3$, are the same as for $SO(3)$ and $\mbs{\rho} \in \mathbb{R}^3$ are the translational parameters (not necessarily equivalent to the actual translation in Euclidean space, $\mbf{r}$).  We would like to generalize the $SO(3)$ result such that we have a surjective-only mapping from $\mathbb{R}^6$ to $SE(3)$.  

Motivated by the Rodrigues-type form of~\eqref{eq:so3map} above, we define a general vector pose mapping to be of the form
\begin{equation}\label{eq:posemap}
\mbf{T}(\mbs{\xi}) = \mbf{1} + a\, \mbs{\xi}^{\wdg} + b\, \mbs{\xi}^{\wdg^2} + c\, \mbs{\xi}^{\wdg^3},
\end{equation}
for some $a$, $b$, and $c$ such that the result is an element of $SE(3)$.  We know that higher-order powers of $\mbs{\xi}^\wdg$ are unnecessary owing to~\eqref{eq:ch2}, which comes from the Cayley-Hamilton theorem.  

It is not immediately obvious that this definition of a pose vector parameterization for $SE(3)$ is consistent with \citet[\S 14.4.1]{bauchau11}, who notes that we can define such a mapping for $\mbox{Ad}(SE(3))$ as one that is {\em parallel to an eigenvector of the resulting adjoint pose matrix associated with the unit eigenvalue}.  It turns out that the two definitions are equivalent and we will demonstrate this shortly after first examining the conditions on $a$, $b$, and $c$ to ensure $\mbf{T} \in SE(3)$.

Using the definition of $\mbs{\xi}^\wdg$, we can manipulate our pose mapping into the form,
\begin{equation}\label{eq:posemap2}
\mbf{T}(\mbs{\xi}) = \bbm \mbf{C}(\mbs{\phi}) & \mbf{D}(\mbs{\phi}) \mbs{\rho} \\ \mbf{0}^T & 1 \ebm = \bbm \mbf{1} + (a - \phi^2 c ) \,\mbs{\phi}^\wdg + b \,\mbs{\phi}^{\wdg^2} & \left( a \mbf{1} + b \,\mbs{\phi}^\wdg + c \,\mbs{\phi}^{\wdg^2} \right) \mbs{\rho} \\ \mbf{0}^T & 1 \ebm,
\end{equation}
where we used~\eqref{eq:ch1} to simplify the top-left entry.

Looking to~\eqref{eq:se3}, the top-left entry in~\eqref{eq:posemap2} must be the rotation matrix:
\begin{equation}\label{eq:Cabc}
\mbf{C}(\mbs{\phi}) = \mbf{1} + (a - \phi^2 c) \,\mbs{\phi}^\wdg + b \,\mbs{\phi}^{\wdg^2}.
\end{equation}
Comparing to~\eqref{eq:so3map}, the following conditions must hold:
\begin{equation}\label{eq:se3mapconditions}
a - \phi^2 c = \frac{\nu^2}{\ep}, \quad b = \frac{\nu^2}{2}.
\end{equation}
The second condition leaves no choice, while the first provides some flexibility.  

Turning to the top-right entry of~\eqref{eq:posemap2}, the matrix
\begin{equation}\label{eq:Dabc}
\mbf{D}(\mbs{\phi}) = a \mbf{1} + b \,\mbs{\phi}^\wdg + c \,\mbs{\phi}^{\wdg^2},
\end{equation}
is the {\em coupling matrix} between the translational and rotational degrees of freedom.  Because there is flexibility in the choice of $a$ and $c$, this means there are an infinite number of possibilities for $\mbf{D}(\mbs{\phi})$ that satisfy our pose mapping definition in~\eqref{eq:posemap}.

Under the conditions in~\eqref{eq:se3mapconditions}, the defined properties of the generating function, and the further condition that $\mbf{D}(\mbs{\phi})$ must be invertible (except at some singularities), we can claim that our chosen mapping in~\eqref{eq:posemap} is surjective-only from $\mathbb{R}^6$ to $SE(3)$.  This is not too hard to see since every $\mbf{C}(\mbs{\phi})$ can be generated by many $\mbs{\phi}$ and every $\mbf{r}$ can be generated by some $\mbs{\rho}$ provided $\mbf{D}(\mbs{\phi})$ invertible.  The inverse map is discussed further in Section~\ref{sec:invmap}.

Returning to the question of consistency of our definition of pose vector parameterizations with that of \citet{bauchau11}, we want to show that $\mbs{\xi}$ {\em is} an eigenvector of $\!\!\Tbig = \mbox{Ad}(\mbf{T})$ associated with a unit eigenvalue, i.e., 
\begin{equation}
\Tbig \mbs{\xi} = \mbs{\xi}.
\end{equation}
Filling in the details of $\!\!\Tbig$ from~\eqref{eq:se3adjointmap1} we have
\begin{equation}
\Tbig \mbs{\xi} = \bbm \mbf{C} & \left( \mbf{D} \mbs{\rho} \right)^\wdg \mbf{C} \\ \mbf{0} & \mbf{C} \ebm \bbm \mbs{\rho} \\ \mbs{\phi} \ebm = \bbm \left( \mbf{C} - \mbs{\phi}^\wdg \mbf{D} \right) \mbs{\rho} \\ \mbs{\phi} \ebm,
\end{equation}
where we have used $\mbf{C} \mbs{\phi} = \mbs{\phi}$.  Using~\eqref{eq:Cabc},~\eqref{eq:Dabc}, and~\eqref{eq:ch1} we see that $\mbf{C} - \mbs{\phi}^\wdg \mbf{D} = \mbf{1}$ so that $\!\!\Tbig \mbs{\xi} = \mbs{\xi}$.  Thus, our $SE(3)$ definition of pose vector parameterizations is equivalent to that of \citet{bauchau11} for $\mbox{Ad}(SE(3))$.  

The nonuniqueness of the pose parameterization that we discussed earlier can also be explained from the eigenstructure, since there are two linearly independent eigenvectors for $\!\!\Tbig$ associated with its two unit eigenvalues \citep{bauchau11,deleuterio_tro21}; the $SE(3)$ pose vector parameterization therefore has a free parameter, as we showed by another means.

\subsection{Screws, Pose Adjoints, and Commutative Mappings}

As discussed in the previous section, we must choose values for the parameters, $a$, $b$, and $c$ to construct the rotation matrix in~\eqref{eq:Cabc}.  However, we have a free parameter in making this choice, as long as the parameters satisfy~\eqref{eq:se3mapconditions}.  One possibility is to choose
\begin{equation}\label{eq:abc}
a = \mu, \quad b = \frac{\nu^2}{2}, \quad c =  \frac{1}{\phi^2}\left( \mu - \frac{\nu^2}{\ep}\right),
\end{equation}
which satisfy the conditions in~\eqref{eq:se3mapconditions} and importantly this particular choice means that we can write our vector pose parameter as \citep[eq. 14.34]{bauchau11}
\begin{equation}\label{eq:screw}
\mbs{\xi} = \phi(\varphi) \, \mbf{s},
\end{equation}
where $\mbf{s}$ is the constant {\em screw axis} of the Mozzi-Chasles theorem.  The screw can be written in the form
\begin{equation}
\mbf{s} = \bbm p \,\mbf{a} + \mbf{m}^\wdg \mbf{a} \\ \mbf{a} \ebm,
\end{equation}
where $p$ is the (scalar) {\em pitch}, $\mbf{a}$ is the (unit) axis of rotation, and $\mbf{m}$ is the moment arm; $(\mbf{m}^\wdg \mbf{a}, \mbf{a})$ are the {\em Pl\"{u}cker coordinates}.  The relationship in~\eqref{eq:screw} parallels the $SO(3)$ definition of \citep{bauchau03} captured in~\eqref{eq:so3def}; we have the product of the generating function and a (screw) axis.  In the case that $\phi(\varphi) = \varphi$, then $\mbs{\xi} = \varphi \mbf{s}$ is the familiar {\em twist}.  We underscore that it is the specific selection of $a$ and $c$ in~\eqref{eq:abc} that has made this connection; other choices for $a$ and $c$ will not result in a constant screw.

The selection of $a$ and $c$ in~\eqref{eq:abc} means that the pose mapping becomes
\begin{equation}\label{eq:se3map2}
\mbf{T}(\mbs{\xi}) = \mbf{1} + \mu  \, \mbs{\xi}^{\wdg} + \frac{\nu^2}{2} \mbs{\xi}^{\wdg^2} + \frac{1}{\phi^2}\left( \mu - \frac{\nu^2}{\ep}\right) \mbs{\xi}^{\wdg^3}.
\end{equation}
As the rotation angle, $\varphi$, becomes small, we have the infinitesimal expression,
\begin{equation}
\mbf{T}(\mbs{\xi}) \approx \mbf{1} + \mbs{\xi}^\wdg,
\end{equation}
which will become important when considering perturbations.

Choosing $a$ and $c$ as in~\eqref{eq:abc} also has the effect of selecting
\begin{equation}
\mbf{D}(\mbs{\phi}) = \mbf{J}(\mbs{\phi}),
\end{equation}
the Jacobian of $SO(3)$ for the chosen parameterization \citep{bauchau11}, which can be verified by substituting into~\eqref{eq:so3jac}.  
Choosing $\mbf{D}(\mbs{\phi}) = \mbf{J}(\mbs{\phi})$ gives our $SE(3)$ mapping an important property related to pose {\em adjoints}.  
With these definitions, we can establish the following desirable commutative relationship:
\begin{equation}
\begin{CD}
4 \times 4@. @. \mbs{\xi}^\wdg  \in  \frak{se}(3) @>\mbox{mapping}>>  \mbf{T} \in SE(3) \\
@. \qquad @. @VV\mbox{ad}V        @VV\mbox{Ad}V \\
6 \times 6 @. @. \mbs{\xi}^\Wdg \in  \mbox{ad}(\frak{se}(3))    @>\mbox{mapping}>>  \Tbig \in \mbox{Ad}(SE(3))
\end{CD}
\end{equation}
This means we can transform our vector parameterization of pose to its adjoint in two equivalent ways.  

To verify this commutative property, we can write the series form for the adjoint mapping as
\begin{equation}\label{eq:se3adjointmap2}
\Tbig(\mbs{\xi}) = \mbf{1} + d \, \mbs{\xi}^\Wdg + e \, \mbs{\xi}^{\Wdg^2} + f \, \mbs{\xi}^{\Wdg^3} + g \, \mbs{\xi}^{\Wdg^4},
\end{equation}
for some unknowns, $d$, $e$, $f$, $g$.  We do not need higher-order terms in this Rodrigues-type series owing to~\eqref{eq:adid}, which can be used to reduce quintic and higher terms.  Multiplying out and comparing terms between~\eqref{eq:se3adjointmap1} and~\eqref{eq:se3adjointmap2}, we find that
\begin{equation}\label{eq:defg}
d = a + \phi^2 (f - c), \quad  e =  \phi^2 g + b,  \quad  f = \frac{1}{2}\left( ab-c \right), \quad g = \frac{1}{2} \left( b^2 - c(a-\phi^2 c) \right),
\end{equation}
in terms of the earlier choices for $a$, $b$, and $c$.  There may also be other choices for $a$ and $c$ (recall these are not unique) that impose the commutative relationship for $SE(3)$, but this was not explored further.

Owing to smoothness of the generating function and its convergence to $\varphi$ in the limit of an infinitesimally small rotation, one might be able to argue that our vector mappings offer alternatives (to the exponential map) for the local diffeomorphism from the Lie algebra, $\frak{se}(3)$, to the Lie group, $SE(3)$.  In such a case, by changing the map we are merely `warping' how the Lie algebra maps onto the Lie group.  

\subsection{Cayley Transformations}

It is known \citep{borri00,bauchau03} that for the Cayley-Gibbs-Rodrigues parameterization of rotation, $\phi(\varphi) = 2 \tan \frac{\varphi}{2}$, we can write the rotation matrix in terms of the Cayley transformation:
\begin{equation}
\mbf{C}(\mbs{\phi}) = \left( \mbf{1} - \frac{1}{2}\mbs{\phi}^\wdg \right)^{-1}  \left( \mbf{1} + \frac{1}{2}\mbs{\phi}^\wdg \right).
\end{equation}
In fact, this can be generalized to the case where $\phi(\varphi) = 2m \tan \frac{\varphi}{2m}$ with $m$ a positive integer using higher-order Cayley transformations \citep{bauchau03}:
\begin{equation}
\mbf{C}(\mbs{\phi}) = \left( \mbf{1} - \frac{1}{2m}\mbs{\phi}^\wdg \right)^{-m}  \left( \mbf{1} + \frac{1}{2m}\mbs{\phi}^\wdg \right)^m.
\end{equation}
However, things become somewhat more complicated for poses.

\citet{borri00} and later \citet{selig07} demonstrated that the Cayley transformation can be used to map pose vectors to $SE(3)$ according to
\begin{equation}\label{eq:se3cay}
\mbf{T}(\mbs{\xi}) = \left( \mbf{1} - \frac{1}{2m}\mbs{\xi}^\wdg \right)^{-m}  \left( \mbf{1} + \frac{1}{2m}\mbs{\xi}^\wdg \right)^m,
\end{equation}
where we have generalized to higher-order transformations.
\citet{borri00} and \citet{selig07} also show that the Cayley transformation can be used to map pose vectors to $\mbox{Ad}(SE(3))$ according to
\begin{equation}\label{eq:se3Adcay}
\Tbig(\mbs{\xi}) = \left( \mbf{1} - \frac{1}{2m}\mbs{\xi}^\Wdg \right)^{-m}  \left( \mbf{1} + \frac{1}{2m}\mbs{\xi}^\Wdg \right)^m,
\end{equation}
where we have generalized again to higher-order transformations.  Disappointingly, \citet{selig07} demonstrates that starting from the same $\mbs{\xi}$ and applying~\eqref{eq:se3cay} and~\eqref{eq:se3Adcay} does not result in an equivalent transformation, i.e., $\Tbig(\mbs{\xi}) \neq \mbox{Ad}(\mbf{T}(\mbs{\xi}))$; the commutative property for adjoints does not hold.

We can actually say something more than \citet{selig07}.  It turns out that~\eqref{eq:se3cay} is {\em not} equivalent to~\eqref{eq:se3map2}, but~\eqref{eq:se3Adcay} {\em is} equivalent to~\eqref{eq:se3adjointmap2}, for the family of rotation parameterizations, $\phi(\varphi) = 2m \tan \frac{\varphi}{2m}$.  The explanation for this lies in the choice of $\mbf{D}(\mbs{\phi})$, discussed earlier.  For the $6 \times 6$ case, choosing $\mbf{D}(\mbs{\phi}) = \mbf{J}(\mbs{\phi})$ results in the Cayley transformation, while for the $4 \times 4$ case it does not; to connect to the Cayley transformation in the $4 \times 4$ case for $m=1$, we must pick $\mbf{D}(\mbs{\phi}) = \frac{1}{2}(\mbf{C}(\mbs{\phi}) + \mbf{1}) \neq \mbf{J}(\mbs{\phi})$, as shown by \citet{selig07}.

To show that~\eqref{eq:se3Adcay} holds, consider the $m=1$ case.  If we left-multiply by the inverse of the first factor we should have
\begin{equation}
\left( \mbf{1} - \frac{1}{2}\mbs{\xi}^\Wdg \right) \Tbig(\mbs{\xi}) =  \mbf{1} + \frac{1}{2}\mbs{\xi}^\Wdg.
\end{equation}
Multiplying out the left side we have
\begin{equation}
\mbf{1} + \left(d-\frac{1}{2}(1-\phi^4 g)\right) \mbs{\xi}^\Wdg + \left(e-\frac{1}{2}d\right) \mbs{\xi}^{\Wdg^2} + \left(f-\frac{1}{2}(e-2\phi^2 g)\right) \mbs{\xi}^{\Wdg^3} + \left(g-\frac{1}{2}f \right) \mbs{\xi}^{\Wdg^4} =  \mbf{1} + \frac{1}{2}\mbs{\xi}^\Wdg,
\end{equation}
where we have made use of~\eqref{eq:adid} once to reduce a quintic term. Comparing the coefficients of each term it must be that
\begin{equation}
d-\frac{1}{2}(1-\phi^4 g) = \frac{1}{2}, \quad e-\frac{1}{2}d = 0, \quad f-\frac{1}{2}(e-2\phi^2 g) = 0, \quad g-\frac{1}{2}f = 0.
\end{equation}
Substituting~\eqref{eq:defg} and then~\eqref{eq:abc} confirms these conditions for $\phi(\varphi) = 2 \tan \frac{\varphi}{2}$.

\subsection{Cayley-Type Factorizations}

Following the previous section that focussed on the $\phi(\varphi) = 2m \tan \frac{\varphi}{2m}$ family, we next consider some novel Cayley-type mappings for different generating functions.  To our knowledge these have not been presented before.

It turns out that we can factor {\em any} of the vectorial rotation parameterizations according to 
\begin{equation}
\mbf{C}(\mbs{\phi}) = \left( \mbf{1} - \lambda \mbs{\phi}^\wdg \right)^{-1} \left( \mbf{1} + \gamma \mbs{\phi}^\wdg \right) = \left( \mbf{1} - \gamma \mbs{\phi}^\wdg \right)^{-1} \left( \mbf{1} + \lambda \mbs{\phi}^\wdg \right),
\end{equation}
where
\beqn{lamgam}
\gamma & = & \frac{1}{2} \ep, \\
\lambda & = & \frac{\nu^2}{\ep} - \frac{1}{2} \ep \left( 1 - \frac{1}{2} \phi^2 \nu^2 \right).
\eeqn
In the case that $\phi(\varphi) = 2 \tan \frac{\varphi}{2}$ we recover $\gamma = \lambda = \frac{1}{2}$.  The inverse of $\mbf{C}(\mbs{\phi})$ is obtained by flipping the signs in the factorization:
\begin{equation}
\mbf{C}(\mbs{\phi})^{-1} = \mbf{C}(\mbs{\phi})^T = \left( \mbf{1} + \lambda \mbs{\phi}^\wdg \right)^{-1} \left( \mbf{1} - \gamma \mbs{\phi}^\wdg \right) = \left( \mbf{1} + \gamma \mbs{\phi}^\wdg \right)^{-1} \left( \mbf{1} - \lambda \mbs{\phi}^\wdg \right).
\end{equation}
The relationships can be verified by simply multiplying out and comparing to~\eqref{eq:so3map}.

We can also factor some $SE(3)$ parameterizations in an analogous way:
\begin{subequations}
\begin{eqnarray}
\mbf{T}(\mbs{\xi}) & = & \left( \mbf{1} - \lambda \mbs{\xi}^\wdg \right)^{-1} \left( \mbf{1} + \gamma \mbs{\xi}^\wdg \right) = \left( \mbf{1} - \gamma \mbs{\xi}^\wdg \right)^{-1} \left( \mbf{1} + \lambda \mbs{\xi}^\wdg \right), \\
\mbf{T}(\mbs{\xi})^{-1} & = & \left( \mbf{1} + \lambda \mbs{\xi}^\wdg \right)^{-1} \left( \mbf{1} - \gamma \mbs{\xi}^\wdg \right) = \left( \mbf{1} + \gamma \mbs{\xi}^\wdg \right)^{-1} \left( \mbf{1} - \lambda \mbs{\xi}^\wdg \right),
\end{eqnarray}
\end{subequations}
where $\gamma$ and $\lambda$ are again given by~\eqref{eq:lamgam}.  To achieve this factorization, we must select $c = \frac{1}{4} \nu^2 \ep$ in~\eqref{eq:se3mapconditions}, which consequently selects a $\mbf{D}(\mbs{\phi}) \neq \mbf{J}(\mbs{\phi})$.

Finally, the pattern holds for $\mbox{Ad}(SE(3))$ where we have
\beqn{}
\Tbig(\mbs{\xi}) & = & \left( \mbf{1} - \lambda \mbs{\xi}^\Wdg \right)^{-1} \left( \mbf{1} + \gamma \mbs{\xi}^\Wdg \right) = \left( \mbf{1} - \gamma \mbs{\xi}^\Wdg \right)^{-1} \left( \mbf{1} + \lambda \mbs{\xi}^\Wdg \right),  \\
\Tbig(\mbs{\xi})^{-1} & = & \left( \mbf{1} + \lambda \mbs{\xi}^\Wdg \right)^{-1} \left( \mbf{1} - \gamma \mbs{\xi}^\Wdg \right) = \left( \mbf{1} + \gamma \mbs{\xi}^\Wdg \right)^{-1} \left( \mbf{1} - \lambda \mbs{\xi}^\Wdg \right),
\eeqn
where $\gamma$ and $\lambda$ are yet again given by~\eqref{eq:lamgam}.  To achieve this, we must select
\begin{equation}
d = \phi^2 f + \frac{\nu^2}{\ep}, \quad e = \frac{1}{2} \ep \phi^2 f + \frac{\nu^2}{2}, \quad f = \frac{\nu^2 \ep}{4 + \phi^2 \ep^2}, \quad g = \frac{1}{2}\ep f,
\end{equation}
in~\eqref{eq:se3adjointmap2}.  As discussed in the previous section, when $\phi(\varphi) = 2 \tan \frac{\varphi}{2}$ this happens to select $\mbf{D}(\mbs{\phi}) = \mbf{J}(\mbs{\phi})$, but this is not true for the other generating functions.

\subsection{Inverse Mappings}\label{sec:invmap}

We briefly discuss how to go the other way from $\mbf{T}$, or $\Tbig$, back to $\mbs{\xi}$ in the general case.  The first step is to extract $\mbf{C}(\mbs{\phi})$ and $\mbf{r}$ from $\mbf{T}$, or $\Tbig$.  From $\mbf{C}(\mbs{\phi})$, we can exploit the facts that
\begin{equation}
\mbox{tr}(\mbf{C}(\mbs{\phi})) = 2 \cos\varphi + 1, \quad \mbf{C}(\mbs{\phi}) \mbf{a} = \mbf{a},
\end{equation}
which are true independent of the chosen vector parameterization.  The first gives us the rotation angle, $\varphi$, while the second is an eigenproblem that can be solved for axis $\mbf{a}$ (unit-length eigenvector corresponding to eigenvalue of $1$);  see \citet{hughes86} for an explicit formula. Owing to the singularity of the vector parameterization, we choose $\varphi \in (-\pi,\pi]$.  We can then calculate $\mbs{\phi} = \phi(\varphi) \, \mbf{a}$.

Once we have $\mbs{\phi}$, we can calculate $\mbs{\rho}$ according to
\begin{equation}
\mbs{\rho} = \mbf{D}(\mbs{\phi})^{-1} \mbf{r},
\end{equation}
and then assemble $\mbs{\rho}$ and $\mbs{\phi}$ into $\mbs{\xi}$.   In the case of $ \mbf{D}(\mbs{\phi}) =  \mbf{J}(\mbs{\phi})$, the inverse, $\mbf{J}(\mbs{\phi})^{-1}$, is given by~\eqref{eq:so3jacinv} and naturally has singularities of which to be aware.

\subsection{Compounding Poses}

For $SO(3)$, we seek to compound two rotations such that
\begin{equation}
\mbf{C}(\mbs{\phi}) = \mbf{C}(\mbs{\phi}_2) \mbf{C}(\mbs{\phi}_1).
\end{equation}
\citet{bauchau03} show that the vector parameterization can be directly compounded according to 
\beqn{}
\varphi & = & 2 \cos^{-1} \left( \nu(\varphi_1)\nu(\varphi_2) \left( \frac{1}{\ep(\varphi_1)\ep(\varphi_2)} - \frac{1}{4} \, \mbs{\phi}_1^T \mbs{\phi}_2 \right)\right), \\
\mbs{\phi} & = & \frac{\nu(\varphi_1) \nu(\varphi_2)}{\nu(\varphi)} \left(  \frac{1}{\ep(\varphi_2)}\mbs{\phi}_1 + \frac{1}{\ep(\varphi_1)}\mbs{\phi}_2 - \frac{1}{2} \mbs{\phi}_1^\wdg \mbs{\phi}_2  \right) .
\eeqn
The first of these is used to calculate the compound rotation angle, $\varphi$, while the second provides the complete vectorial parameterization, $\mbs{\phi}$.

For $SE(3)$, \citet{condurache20} recently discussed a closed-form solution for compounding two pose vectors in the case that $\mbs{\phi} = \varphi \mbf{a}$.  \citet{bauchau03b} showed that we can easily generalize this result to any of our vectorial pose parameterizations.  The compounding of two poses is 
\begin{multline}
\Tsmall(\mbs{\xi}) = \bbm \mbf{C}(\mbs{\phi}) & \mbf{D}(\mbs{\phi}) \mbs{\rho} \\ \mbf{0}^T & 1 \ebm = \Tsmall(\mbs{\xi}_2) \Tsmall(\mbs{\xi}_1) = \bbm \mbf{C}(\mbs{\phi}_2) & \mbf{D}(\mbs{\phi}_2) \mbs{\rho}_2 \\ \mbf{0}^T & 1 \ebm \bbm \mbf{C}(\mbs{\phi}_1) & \mbf{D}(\mbs{\phi}_1) \mbs{\rho}_1 \\ \mbf{0}^T & 1 \ebm \\ = \bbm \mbf{C}(\mbs{\phi}_2) \mbf{C}(\mbs{\phi}_1) & \mbf{C}(\mbs{\phi}_2) \mbf{D}(\mbs{\phi}_1)\mbs{\rho}_1  + \mbf{D}(\mbs{\phi}_2) \mbs{\rho}_2 \\ \mbf{0}^T & 1 \ebm .
\end{multline}
Comparing entries we see that
\beqn{}
\mbf{C}(\mbs{\phi}) & =&  \mbf{C}(\mbs{\phi}_2) \mbf{C}(\mbs{\phi}_1), \\
\mbs{\rho} & = & \mbf{D}(\mbs{\phi})^{-1} \mbf{C}(\mbs{\phi}_2) \mbf{D}(\mbs{\phi}_1)\mbs{\rho}_1 + \mbf{D}(\mbs{\phi})^{-1} \mbf{D}(\mbs{\phi}_2) \mbs{\rho}_2 . 
\eeqn
We can use the approach of \citet{bauchau03} to calculate the compound rotation, $\mbs{\phi}$, and then calculate $\mbs{\rho}$ afterwards.  We require an expression for $ \mbf{D}(\mbs{\phi})^{-1}$, which in the case of $ \mbf{D}(\mbs{\phi}) =  \mbf{J}(\mbs{\phi})$ is given by~\eqref{eq:so3jacinv}.

\subsection{Additional Useful Properties of  Vector Mappings}

There are a few other important properties of our vector mappings of which we frequently make use when manipulating expressions in state-estimation problems.  Owing to the well-known identity \citep{barfoot_ser17},
\begin{equation}
\left( \mbf{C} \mbf{v} \right)^\wdg \equiv \mbf{C} \mbf{v}^\wdg \mbf{C}^T,
\end{equation}
where $\mbf{C} \in SO(3)$ and $\mbf{v} \in \mathbb{R}^3$, we have that
\begin{equation}
\mbf{C}_1 \mbf{C}_2(\mbs{\phi}_2) \mbf{C}_1^T \equiv \mbf{C}_2\left( \mbf{C}_1 \mbs{\phi}_2 \right).
\end{equation}
The proof is as follows:
\begin{multline}\label{eq:Cv}
\mbf{C}_1 \mbf{C}_2(\mbs{\phi}_2) \mbf{C}_1^T = \mbf{C}_1 \left(  \mbf{1} + \alpha \, \mbs{\phi}_2^\wdg + \beta \mbs{\phi}_2^{\wdg^2}  \right) \mbf{C}_1^T = \mbf{1} + \alpha \, \mbf{C}_1 \mbs{\phi}_2^\wdg  \mbf{C}_1^T + \beta \, \mbf{C}_1 \mbs{\phi}_2^\wdg \mbf{C}_1 \, \mbf{C}_1^T \mbs{\phi}_2^\wdg \mbf{C}_1^T \\
= \mbf{1} + \alpha \left( \mbf{C}_1 \mbs{\phi}_2 \right)^\wdg + \beta \left( \mbf{C}_1 \mbs{\phi}_2 \right)^{\wdg^2} = \mbf{C}_2\left( \mbf{C}_1 \mbs{\phi}_2 \right),
\end{multline}
where $\alpha$ and $\beta$ are standing in for the coefficients in~\eqref{eq:so3map}.

For $SE(3)$, we can use the identities \citep{barfoot_ser17},
\begin{gather}
\left( \!\! \Tbig \mbf{x} \right)^\wdg \equiv \mbf{T} \mbf{x}^\wdg \mbf{T}^{-1}, \\ \left( \!\!\Tbig \mbf{x} \right)^\Wdg \equiv \Tbig \mbf{x}^\Wdg \Tbig^{-1}
\end{gather}
with $\mbf{T} \in SE(3)$, $\Tbig = \mbox{Ad}(\mbf{T})$, $\mbf{x} \in \mathbb{R}^6$ to establish
\begin{gather}
\mbf{T}_1 \mbf{T}_2(\mbs{\xi}_2) \mbf{T}_1^{-1} \equiv \mbf{T}_2 \left( \!\!\Tbig_1 \mbs{\xi}_2 \right), \\ \Tbig_1 \Tbig_2(\mbs{\xi}_2) \Tbig_1^{-1} \equiv \Tbig_2 \left( \!\!\Tbig_1 \mbs{\xi}_2 \right),
\end{gather}
where the proofs are similar to~\eqref{eq:Cv}.

\section{Applications}\label{sec:apps}

In this section, we touch on three classic applications of pose vectors:  (i) interpolation between two poses (useful in computer graphics, for example), (ii) a servoing control problem, and (iii) the alignment of two pointclouds.  In each, we show how different choices for the vector parameterization can produce different results.

\subsection{Interpolation}

\begin{figure*}[t]
\centering
\includegraphics[width=0.95\textwidth]{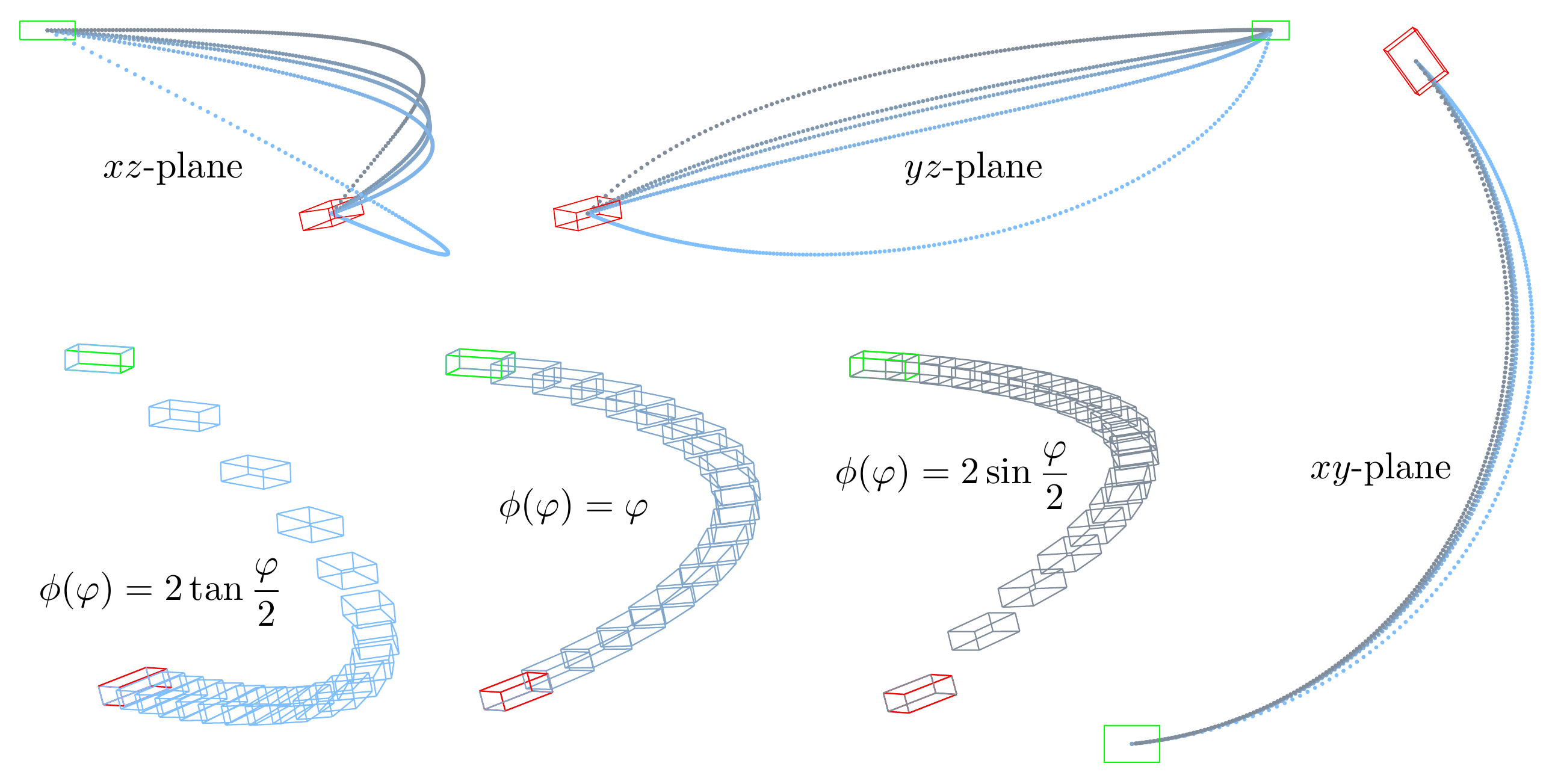}
\caption{An example of  `linear' interpolation between an initial pose (green) and final pose (red) for different vector parameterizations of pose.  The paths are straight lines in $\mathbb{R}^6$ but when mapped to $SE(3)$ produce different curves.  Each parameterization takes a unique path with the $\phi(\varphi) = 2\tan\frac{\varphi}{2}$ generating function being the most distinct.  The figure compares the path taken by each of the generating functions from Figure~\ref{fig:genfunc} (same colour scheme) as well as displays the full pose trajectory individually for three generating functions}
\label{fig:example1}
\end{figure*}

Pose interpolation has been discussed extensively in the literature; \citet{han16,han18} provide some representative examples.  Here our goal is not to claim a new method of pose interpolation but rather to show the effect of simply changing the generating function on a basic interpolation scheme defined in the vectorspace.

We can define a `linear' interpolation scheme for poses using any of our vector parameterizations:
\begin{equation}
\mbs{\xi} = (1-s) \, \mbs{\xi}_{\rm final} + s \, \mbs{\xi}_{\rm initial},
\end{equation}
where $\mbs{\xi}_{\rm initial}$ is an initial pose, $\mbs{\xi}_{\rm final}$ a final pose, and $s \in [0,1]$ the interpolation variable.  As we vary $s$ from $0$ to $1$, the vector, $\mbs{\xi}$ will vary and so too will $\mbf{T}(\mbs{\xi})$.  This allows the smooth transition from one pose to another.

Figure~\ref{fig:example1} provides a qualitative example of this scheme for all five generating functions, $\phi(\varphi)$, listed in Table~\ref{tab:so3vecparams} with $\mbf{D}(\mbs{\phi}) = \mbf{J}(\mbs{\phi})$.  The paths taken each correspond to a {\em geodesic} in $\mathbb{R}^6$, but when mapped to $SE(3)$ are quite different.  Notably, none of the paths are straight lines in Euclidean space owing to the coupling between translation and rotation.  As the rotational motion becomes smaller, the paths followed by the different generating functions look more similar owing to~\eqref{eq:genprop}.  Looking more closely at Figure~\ref{fig:example1}, we can also observe that the spacing of the steps are quite different for the different generating functions, even though we stepped $s$ uniformly.  The $\tan$ generating functions take larger steps at the beginning but smaller ones at the end, while the $\sin$ ones do the opposite, and the $\phi(\varphi) = \varphi$ function keeps the steps even throughout.  Naturally, we can make the steps in the interpolation variable, $s$, as small as we like to obtain smooth interpolation; we chose fairly large steps to keep the plot in Figure~\ref{fig:example1} from becoming too cluttered.

\subsection{Pose Servoing Control}

\begin{figure*}[t]
\centering
\includegraphics[width=0.75\textwidth]{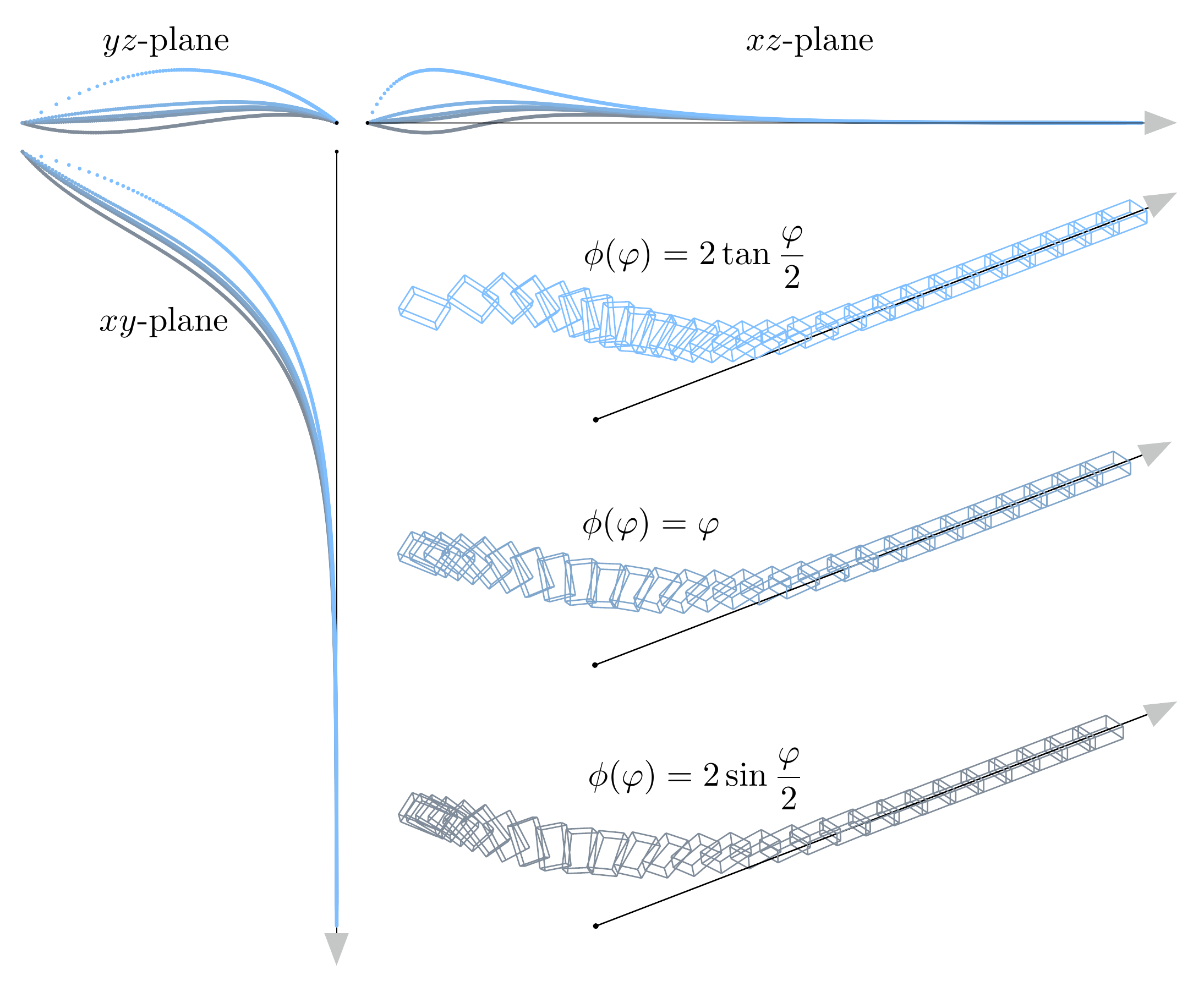}
\caption{An example of pose servoing control using a `linear' controller in pose vector space.  Once mapped to $SE(3)$ we see different behaviours from the different pose mappings.  In this example, the reference pose is moving along the straight black line and there is an initial error.  The figure compares the path taken to join the reference by each of the generating functions from Figure~\ref{fig:genfunc} (same colour scheme) as well as displays the full pose trajectory individually for three generating functions}
\label{fig:example2}
\end{figure*}

Pose servoing is also a well studied topic. Here we simply want to show the effect of changing the pose mapping on a simple control law defined in the vectorspace.  We define a `linear' control law to servo onto a reference path.  Consider a reference pose, $\mbf{T}_{\rm ref}$, that we want to track.  We can compare our current pose, $\mbf{T}(\mbs{\xi})$, to this reference and put it through the inverse mapping to create an `error vector':
\begin{equation}
\mbs{\xi}_{\rm error} \leftarrow \mbf{T}(\mbs{\xi}) \mbf{T}_{\rm ref}^{-1},
\end{equation}
where the left arrow stands in for the inverse of any of our pose mappings.  We can then adjust our pose to be slightly closer to the reference pose (which is moving) using a simple control law:
\begin{equation}
\dot{\mbs{\xi}} = -\kappa \, \mbs{\xi}_{\rm error}.
\end{equation}
with $\kappa >0$ a gain.  As the error $\mbs{\xi}_{\rm error}$ approaches zero, the moving pose, $\mbf{T}(\mbs{\xi})$ will approach the reference pose, $\mbf{T}_{\rm ref}$.

Figure~\ref{fig:example2} provides an example of servoing to a reference pose that is moving straight along the $x$-axis, starting from a poor initial pose that is translated and rotated from the reference.  We kept the vector control law (including the gain) identical and varied only the choice of pose mapping.  As expected, we see different transient behaviour resulting from the different pose mappings as they seek the reference.

\subsection{Pointcloud Alignment}

We use this section to discuss aligning two clouds of three-dimensional points.  Our problem is similar to that of \citet{barfoot_aa11}, but formulated directly on $SE(3)$.

\subsubsection{Problem Setup}

Let us consider that we have two sets of observations of some three-dimensional points, $\mbf{p}_j$ and $\mbf{q}_j$, expressed in two different reference frames.  We assume the points are expressed in homogeneous format,
\begin{equation}
\mbf{p} = \bbm \mbf{v} \\ 1 \ebm, \quad \mbf{v} = \bbm x \\ y \\ z \ebm .
\end{equation}
Our goal is to find the rigid-body transformation, $\mbf{T}$,  that minimizes the following sum-of-squares cost function,
\begin{equation}
J(\mbf{T}) = \frac{1}{2} \sum_j \mbf{e}_j^T \mbf{W}_j \mbf{e}_j,
\end{equation}
where
\begin{equation}
\mbf{e}_j = \mbf{q}_j - \mbf{T} \, \mbf{p}_j,
\end{equation}
and $\mbf{W}_j > 0$ is a matrix weight for each associated point pair.

\subsubsection{General Perturbation Solution}\label{sec:genper}

Owing to the matrix weights, there is no closed-form solution to this optimization problem and we must turn to an iterative scheme \citep{barfoot_ser17}.  If we have some initial guess (or operating point) for our pose solution, $\mbf{T}_{\rm op}$, we can seek an optimal perturbation, $\delta\mbf{T}(\mbs{\xi})$, that iteratively reduces our cost function.  Any of the pose mappings discussed earlier could be used in this regard:
\begin{equation}
\mbf{T} = \delta\mbf{T}(\mbs{\xi}) \, \mbf{T}_{\rm op} \approx \left( \mbf{1} + \mbs{\xi}^\wdg \right) \mbf{T}_{\rm op},
\end{equation}
where notably the infinitesimal expression is the same for all our pose mappings.
Inserting this into the error we have
\begin{equation}
\mbf{e}_j \approx \mbf{q}_j - \left( \mbf{1} + \mbs{\xi}^\wdg \right) \mbf{T}_{\rm op} \mbf{p}_j = \underbrace{\mbf{q}_j - \mbf{T}_{\rm op} \mbf{p}_j}_{\mbf{e}_{j,{\rm op}}} - \left( \mbf{T}_{\rm op} \mbf{p}_j \right)^\pfs \mbs{\xi} = \mbf{e}_{j,{\rm op}} - \left( \mbf{T}_{\rm op} \mbf{p}_j \right)^\pfs \mbs{\xi},
\end{equation}
where we have made use of $\mbs{\xi}^\wdg \mbf{p} \equiv \mbf{p}^\pfs \mbs{\xi}$ \citep{barfoot_tro14, barfoot_ser17} with
\begin{equation}
\mbf{p}^\pfs = \bbm \mbf{1} & -\mbf{v}^\wdg \\ \mbf{0}^T & \mbf{0}^T \ebm. 
\end{equation}
Our approximate error expression is now linear and inserting it into the cost function we have
\begin{equation}\label{eq:cost1}
J(\mbs{\xi}) \approx \frac{1}{2} \sum_j \left( \mbf{e}_{j,{\rm op}} - \left( \mbf{T}_{\rm op} \mbf{p}_j \right)^\pfs \mbs{\xi} \right)^T \mbf{W}_j \left( \mbf{e}_{j,{\rm op}} - \left( \mbf{T}_{\rm op} \mbf{p}_j \right)^\pfs \mbs{\xi} \right),
\end{equation}
which is exactly quadratic in $\mbs{\xi}$, and can be minimized by solving the following linear system of equations for $\mbs{\xi}$:
\begin{equation}
\left(\sum_j \left( \mbf{T}_{\rm op} \mbf{p}_j \right)^{\pfs^T} \mbf{W}_j  \left( \mbf{T}_{\rm op} \mbf{p}_j \right)^\pfs  \right) \, \mbs{\xi} = \sum_j \left( \mbf{T}_{\rm op} \mbf{p}_j \right)^{\pfs^T} \mbf{W}_j \mbf{e}_{j,{\rm op}}.
\end{equation}
After solving for the optimal $\mbs{\xi}$, we can update our initial guess using our pose mapping of choice,
\begin{equation}
\mbf{T}_{\rm op} \leftarrow \delta\mbf{T}(\mbs{\xi}) \, \mbf{T}_{\rm op},
\end{equation}
and iterate to convergence.  This approach is discussed in detail by \citet{barfoot_ser17} for the $\phi(\varphi) = \varphi$ case, while here we show it generalizes to any vectorial mapping of pose.

\subsubsection{Alternate Cayley Perturbation Solution}\label{sec:cayper}

The previous section relied on using the infinitesimal expression for pose at each iteration.  Here we take a slightly different strategy based on the Cayley transformation that is similar in spirit to that of \citep{mortari07,majji11,junkins11,wong16,wong18,qian20}.  We will refer to the resulting algorithm as {\em CayPer}.  If we choose the $\phi(\varphi) = 2 \tan \frac{\varphi}{2}$ generating function we can then let our perturbation to the operating point be
\begin{equation}
\delta \mbf{T}(\mbs{\xi}) = \left( \mbf{1} - \frac{1}{2} \mbs{\xi}^\wdg \right)^{-1}  \left( \mbf{1} + \frac{1}{2} \mbs{\xi}^\wdg \right).
\end{equation}
Our point error is then
\begin{equation}
\mbf{e}_j = \mbf{q}_j - \left( \mbf{1} - \frac{1}{2} \mbs{\xi}^\wdg \right)^{-1}  \left( \mbf{1} + \frac{1}{2} \mbs{\xi}^\wdg \right) \mbf{T}_{\rm op} \, \mbf{p}_j.
\end{equation}
We can then define a new error, $\mbf{e}_j^\prime$, by premultiplying by $ \left( \mbf{1} - \frac{1}{2} \mbs{\xi}^\wdg \right)$:
\begin{equation}
\mbf{e}_j^\prime = \left( \mbf{1} - \frac{1}{2} \mbs{\xi}^\wdg \right) \mbf{e}_j = \left( \mbf{1} - \frac{1}{2} \mbs{\xi}^\wdg \right) \mbf{q}_j - \left( \mbf{1} + \frac{1}{2} \mbs{\xi}^\wdg \right) \mbf{T}_{\rm op} \mbf{p}_j = \mbf{e}_{j,{\rm op}} - \frac{1}{2}\left( \mbf{q}_j + \mbf{T}_{\rm op} \mbf{p}_j  \right)^\pfs \, \mbs{\xi}.
\end{equation}
Importantly, we have that
\begin{equation}
\lim_{\mbs{\xi} \rightarrow \mbf{0}} \mbf{e}^\prime_j = \mbf{e}_j.
\end{equation}
This suggests that we can define a new cost function,
\begin{multline}
J^\prime(\mbs{\xi}) = \frac{1}{2} \sum_j \mbf{e}_j^{\prime^T} \mbf{W}_j \mbf{e}_j^\prime \\ \approx \frac{1}{2} \sum_j \left( \mbf{e}_{j,{\rm op}} - \frac{1}{2}\left( \mbf{q}_j + \mbf{T}_{\rm op} \mbf{p}_j  \right)^\pfs \, \mbs{\xi} \right)^T \mbf{W}_j \left( \mbf{e}_{j,{\rm op}} - \frac{1}{2}\left( \mbf{q}_j + \mbf{T}_{\rm op} \mbf{p}_j  \right)^\pfs \, \mbs{\xi} \right),
\end{multline}
which we minimize at each iteration instead of~\eqref{eq:cost1}.  The optimal perturbation, $\frac{1}{2}\mbs{\xi}$, is this time the solution to the linear system,
\begin{equation}
\left(\sum_j \left( \mbf{q}_j + \mbf{T}_{\rm op} \mbf{p}_j \right)^{\pfs^T} \mbf{W}_j  \left(  \mbf{q}_j + \mbf{T}_{\rm op} \mbf{p}_j \right)^\pfs  \right) \, \frac{1}{2} \mbs{\xi} = \sum_j \left( \mbf{q}_j + \mbf{T}_{\rm op} \mbf{p}_j \right)^{\pfs^T} \mbf{W}_j \mbf{e}_{j,{\rm op}}.
\end{equation}
The update to the operating point is then
\begin{equation}
\mbf{T}_{\rm op} \leftarrow  \left( \mbf{1} - \frac{1}{2} \mbs{\xi}^\wdg \right)^{-1}  \left( \mbf{1} + \frac{1}{2} \mbs{\xi}^\wdg \right) \, \mbf{T}_{\rm op},
\end{equation}
which requires only linear algebraic operations (e.g., no trigonometric functions).  Notably, on the first iteration with $\mbf{T}_{\rm op} = \mbf{1}$, the CayPer approach becomes a one-shot Weighted Optimal Linear Attitude and Translation Estimator (WOLATE) \citep{mortari07,majji11,junkins11,wong16,wong18,qian20} directly on $SE(3)$.  However, we can also continue to iterate to convergence of the original problem. 

\subsubsection{Experiment}

We conducted a simulated pointcloud alignment experiment motivated by aligning two sets of matched features from a stereo camera.  Our approach is very similar in nature to that of \citet{maimone07} and \citet{barfoot_aa11} in that we align matched stereo points in Euclidean space rather than image space.  

We generated feature locations in image space, $\mbf{y}$, by observing a set of static landmarks using a stereo camera model:
\begin{equation}
\mbf{y} = \bbm u_\ell \\ v_\ell \\ u_r \\ v_r \ebm  = \underbrace{\bbm f & 0 & c_u & f \frac{b}{2} \\ 0 & f & c_v & 0 \\ f & 0 & c_u & -f\frac{b}{2} \\ 0 & f & c_v & 0 \ebm}_{\mbf{M}} \, \frac{1}{z} \bbm x \\ y \\ z \\ 1 \ebm + \mbf{n}, 
\end{equation}
where $\mbf{n} \sim \mathcal{N}(\mbf{0}, \mbf{R})$ and $\mbf{M}$ is a combined parameter matrix for the stereo rig.  We used focal length $f = 200$ [pixels], stereo baseline $b = 0.25$ [m], optical center $(c_u, c_v) = (0,0)$ [pixels], and measurement covariance $\mbf{R} = 0.25^2 \mbf{1}$ [pixels$^2$].  This model was used to generate two noisy pointclouds where the second frame is transformed by groundtruth pose, $\mbf{T}_{\rm gt}$, from the first.  Once the noisy feature locations were generated in the two different stereo image frames, we transformed these back to two Euclidean pointclouds, $\mbf{p}_j$ and $\mbf{q}_j$, using the inverse stereo camera model.  The alignment problem is then to recover $\mbf{T}_{\rm gt}$ using $\mbf{p}_j$ and $\mbf{q}_j$.  We assume the points have already been correlated correctly and do not treat the outlier rejection problem here.

\begin{figure*}[t]
\centering
\includegraphics[width=0.49\textwidth]{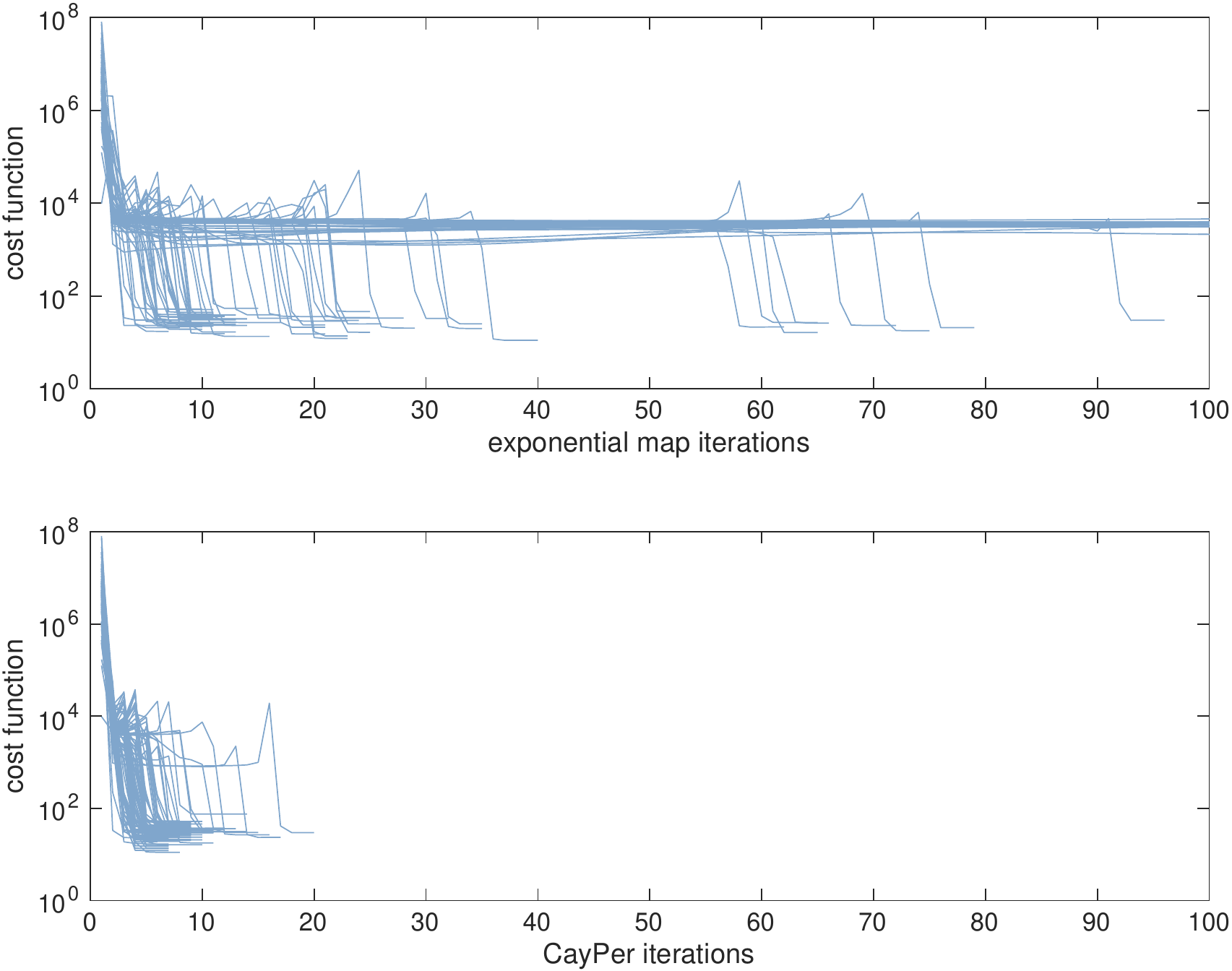}
\includegraphics[width=0.49\textwidth]{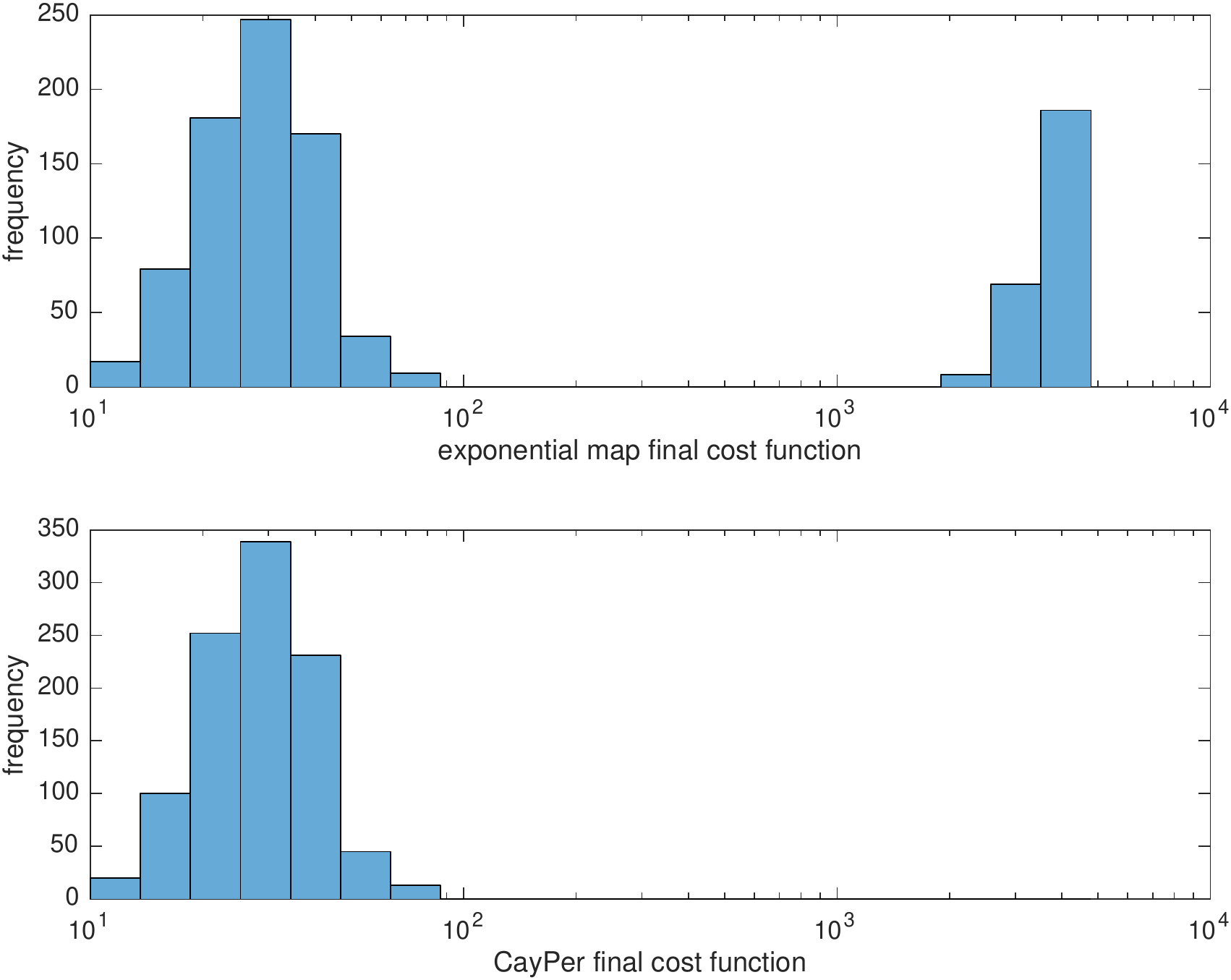}
\caption{Results of pointloud alignment problem for both the exponential map and CayPer algorithm.  The left plot shows the convergence history of the cost function $J$ for $100$ trials where the initial pose was randomly selected.  We see that the exponential map regularly takes a long time to converge and often becomes stuck in a local minimum when the initial pose is far from the groundtruth.  CayPer always converged in $20$ iterations or less and always ended at the global minimum.  The right plot shows a histogram of the costs, $J$, at the final iteration for $1000$ trials where the initial pose was randomly selected.  Again, we see that in approximately $20$\% of the trials the exponential map failed to converge to the global minimum while CayPer succeeded in all trials}
\label{fig:example3}
\end{figure*}

To generate the weight matrices, $\mbf{W}_j$, for our cost function, we mapped the isotropic covariance in image space, $\mbf{R}$, through the inverse stereo camera model to produce an (inverse) covariance of our error in Euclidean space:
\begin{equation}
\mbf{W}_j^{-1} = \mathbb{E} [ \mbf{e}_j \mbf{e}_j^T ].
\end{equation}
The resulting $\mbf{W}_j^{-1}$ is a long, skinny uncertainty ellipsoid in the depth direction of the camera as established by \citet{matthies87}.  This anisotropic measurement covariance makes this problem particularly challenging.  As far we know, there is no practical algorithm that can guarantee finding the global minimum of the cost function, $J$, in this anisotropic situation.  \citet{yang20} show how to certify finding the global minimum in the case that $\mbf{W}_j^{-1}$ is isotropic (i.e., a scalar times the identity matrix) even with significant outliers, but this does not apply in the chosen situation.

We selected a pyramid distribution of $12$ landmarks in the field of view from $5$ to $15$ m away from the camera.  The camera was translated $1$ m in the depth, $z$, direction to generate the two noisy pointclouds.  We then ran many trials where we varied (i) the random draw of the measurement noise, and (ii) the random draw of the initial guess for the alignment algorithms.  Our goal was to see if choosing different vector parameterizations of pose would result in different paths to get from the initial guess to the groundtruth pose. 

\subsubsection{Results}

\begin{figure*}[p]
\centering
\includegraphics[width=0.75\textwidth]{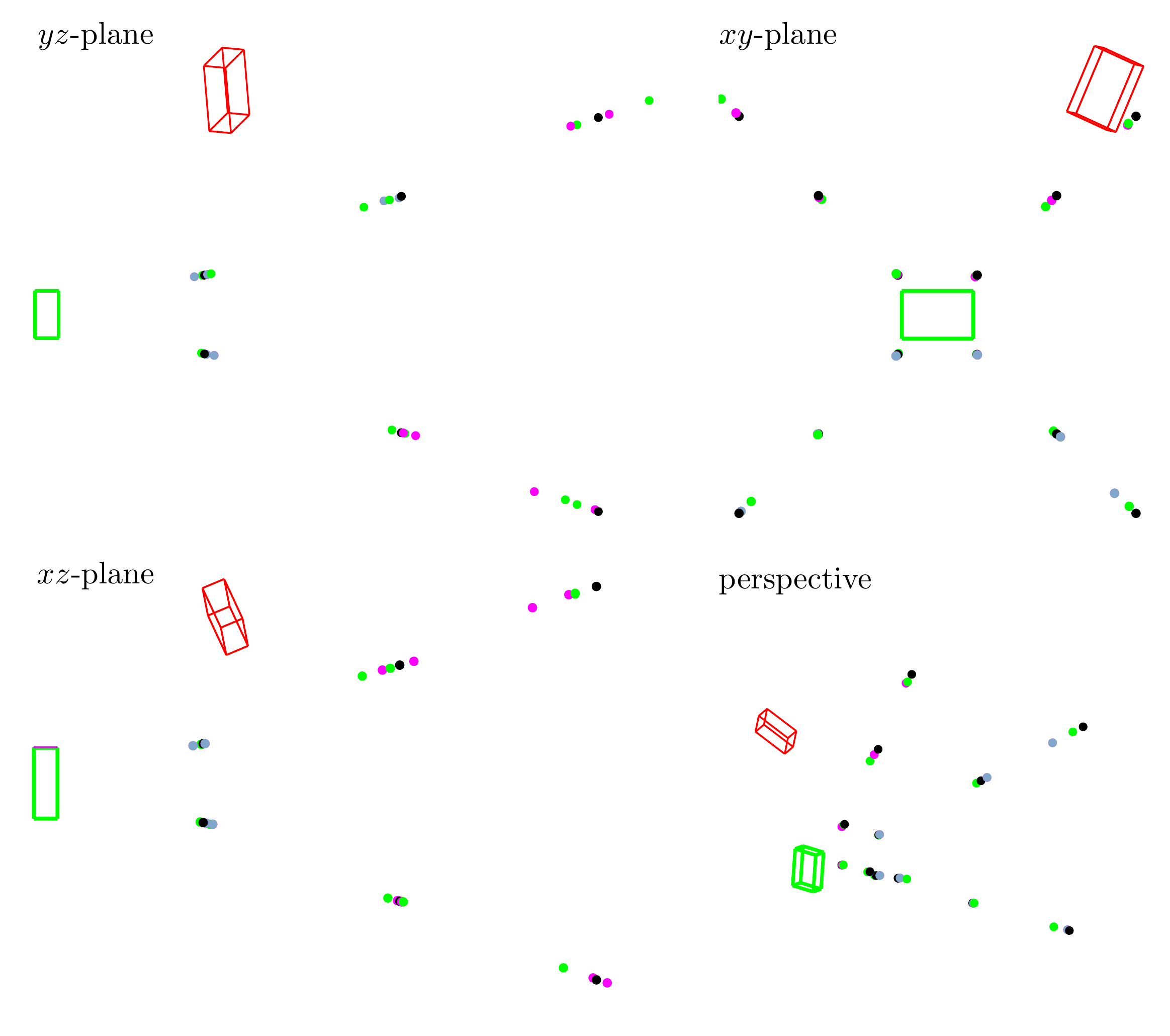}
\rule{0.7\textwidth}{0.4pt}
\includegraphics[width=0.75\textwidth]{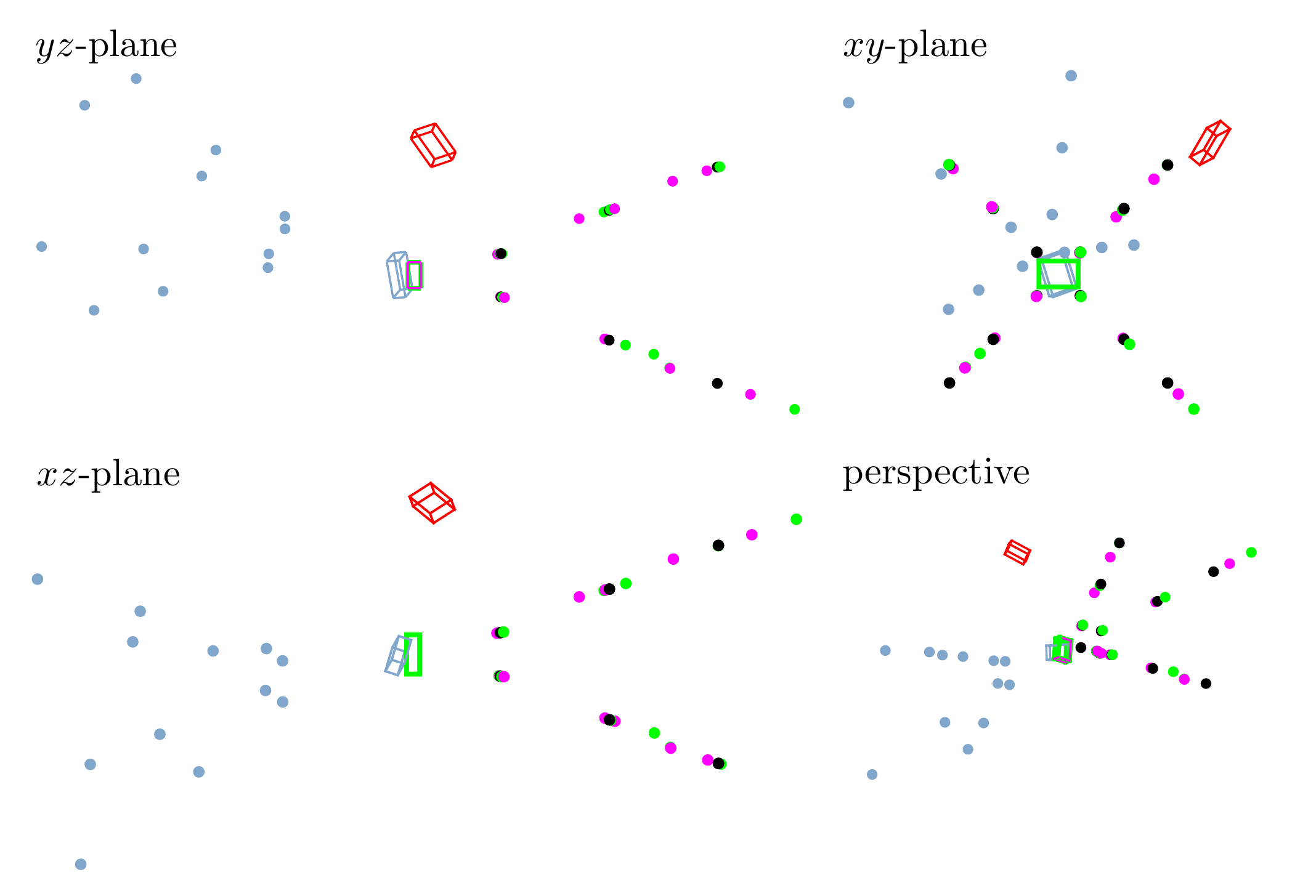}
\caption{Two trials of the pointcloud alignment problem.  In both cases, CayPer successfully finds the correct minimum while the exponential map succeeds in the top case but not the bottom where it becomes stuck in an undesirable local minimum.  Red box is the initial guess, green box is the groundtruth pose, black dots are the groundtruth landmark locations, green dots are the first noisy pointcloud (static), magenta box (hidden by green box) is the CayPer final pose, blue box (when not hidden by green box) is the exponential map final pose, magenta dots are the second noisy pointcloud (transformed) at the final CayPer pose, blue dots are the second noisy pointlcoud (transformed) at the final exponential map pose}
\label{fig:example3b}
\end{figure*}

We found that using different vector parameterizations of pose in the general perturbation approach of Section~\ref{sec:genper} made little difference and so selected the `exponential map' as a representative method.  However, the CayPer approach of Section~\ref{sec:cayper} turned out to be a lot better at not getting stuck in poor local minima.

Figure~\ref{fig:example3} shows quantitative convergence results for several trials.  We see that the CayPer approach does not get stuck in local minima as easily as the exponential map approach.  In fact, in $1000$ trials CayPer always arrived at the global minimum, but we are not prepared to claim this will always be true at this point.   The exponential map frequently became trapped in local minima when the initial guess was particularly bad.  We also found that even when both algorithms did find the global minimum, CayPer took fewer iterations to get there; both algorithms were iterated until $\mbs{\xi}^T \mbs{\xi} < 10^{-10}$ or hit $100$ iterations.

Figure~\ref{fig:example3b} shows two of the $1000$ trials we ran.  The top case shows the final alignment for a trial where both algorithms successfully converged to the global minimum.  The bottom shows a trial where CayPer succeeded but the exponential map failed (catastrophically).

\section{Conclusion and Future Work}\label{sec:confu}

We presented a family of different vector mappings of pose that may find application in computer vision, robotics, and even graphics.  We showed that given a vector parameterization of rotation, there is actually an infinite number of possible vector pose mappings with different couplings between the rotational and translational components of the pose.   We explored commutative mappings as well as Cayley and Cayley-type mappings.  We discussed inverse mappings as well as provided a general formula for the compounding of two general pose mappings.  Finally, we showed several of these pose mappings in action on three different applications:  pose interpolation, pose servoing control, and pointcloud alignment estimation.  The main conclusion is that while we might think of the exponential map as the {\em canonical} way to map vectors to poses, there are many other possibilities that might be preferable, depending on the application.  In particular, we found that the Cayley transformation was particularly successful in the CayPer pointcloud alignment method we presented.

Moving forward, there are certainly other theoretical results and connections that might be found related to vector pose mappings.  Moreover, several of these ideas might extend to analysis of other useful matrix Lie groups that involve $SO(3)$ as a building block.  On the practical side, further analysis of the CayPer algorithm is worthwhile.  We also believe that using the Cayley transformation to iteratively optimize poses in more involved problems such as pose-graph optimization and simultaneous localization and mapping will be worth pursuing.

%

\vspace*{-0.1in}
\section*{Acknowledgement}

This work was partially supported by the Natural Sciences and Engineering Research Council (NSERC) of Canada.

%
%
%

\vspace*{-0.1in}
\bibliographystyle{asrl}
\bibliography{refs}

\begin{thebibliography}{30}
\newcommand{\enquote}[1]{``#1''}
\providecommand{\natexlab}[1]{#1}

\bibitem[{Ball(1900)}]{ball1900}
Ball, R.~S., \emph{A Treatise on the Theory of Screws}, Cambridge university
  press, 1900.

\bibitem[{Barfoot(2017)}]{barfoot_ser17}
Barfoot, T.~D., \emph{State Estimation for Robotics}, Cambridge University
  Press, 2017.

\bibitem[{Barfoot et~al.(2011)Barfoot, Forbes, and Furgale}]{barfoot_aa11}
Barfoot, T.~D., Forbes, J.~R., and Furgale, P.~T., \enquote{Pose Estimation
  using Linearized Rotations and Quaternion Algebra,} \emph{Acta Astronautica},
  68(1-2):101--112, 2011.

\bibitem[{Barfoot and Furgale(2014)}]{barfoot_tro14}
Barfoot, T.~D. and Furgale, P.~T., \enquote{Associating Uncertainty with
  Three-Dimensional Poses for use in Estimation Problems,} \emph{IEEE
  Transactions on Robotics}, 30(3):679--693, 2014,
  (\href{http://asrl.utias.utoronto.ca/code/}{code}).

\bibitem[{Bauchau(2011)}]{bauchau11}
Bauchau, O.~A., \emph{Flexible multibody dynamics}, Springer Science \&
  Business Media, 2011.

\bibitem[{Bauchau and Choi(2003)}]{bauchau03b}
Bauchau, O.~A. and Choi, J.-Y., \enquote{The vector parameterization of
  motion,} \emph{Nonlinear Dynamics}, 33(2):165--188, 2003.

\bibitem[{Bauchau and Li(2011)}]{bauchau11b}
Bauchau, O.~A. and Li, L., \enquote{Tensorial parameterization of rotation and
  motion,} \emph{Journal of computational and nonlinear dynamics}, 6(3), 2011.

\bibitem[{Bauchau and Trainelli(2003)}]{bauchau03}
Bauchau, O.~A. and Trainelli, L., \enquote{The vectorial parameterization of
  rotation,} \emph{Nonlinear dynamics}, 32(1):71--92, 2003.

\bibitem[{Borri et~al.(2000)Borri, Trainelli, and Bottasso}]{borri00}
Borri, M., Trainelli, L., and Bottasso, C.~L., \enquote{On representations and
  parameterizations of motion,} \emph{Multibody System Dynamics},
  4(2):129--193, 2000.

\bibitem[{Cayley(1846)}]{cayley1846}
Cayley, A., \enquote{Sur quelques propri{\'e}t{\'e}s des d{\'e}terminants
  gauches.} \emph{Journal f{\"u}r die reine und angewandte Mathematik},
  1846(32):119--123, 1846.

\bibitem[{Chasles(1830)}]{chasles1830}
Chasles, M., \enquote{Note sur les propri{\'e}t{\'e}s g{\'e}n{\'e}rales du
  syst{\`e}me de deux corps semblables entr'eux et plac{\'e}s d'une mani{\`e}re
  quelconque dans l'espace; et sur le d{\'e}placement fini ou infiniment petit
  d'un corps solide libre,} \emph{Bulletin des Sciences Math{\'e}matiques,
  F{\'e}russac}, 14:321--26, 1830.

\bibitem[{Chirikjian(2009)}]{chirikjian09}
Chirikjian, G.~S., \emph{Stochastic Models, Information Theory, and {Lie}
  Groups: Classical Results and Geometric Methods}, volume 1-2, Birkhauser, New
  York, 2009.

\bibitem[{Condurache and Ciureanu(2020)}]{condurache20}
Condurache, D. and Ciureanu, I.-A., \enquote{Baker--Campbell--Hausdorff--Dynkin
  Formula for the Lie Algebra of Rigid Body Displacements,} \emph{Mathematics},
  8(7):1185, 2020.

\bibitem[{D'Eleuterio and Barfoot(2021)}]{deleuterio_tro21}
D'Eleuterio, G. M.~T. and Barfoot, T.~D., \enquote{On the Eigenstructure of
  Rotations and Poses: Commonalities and Peculiarities,} 2021, in preparation
  for submission.

\bibitem[{Han and Bauchau(2016)}]{han16}
Han, S. and Bauchau, O.~A., \enquote{Manipulation of motion via dual entities,}
  \emph{Nonlinear Dynamics}, 85(1):509--524, 2016.

\bibitem[{Han and Bauchau(2018)}]{han18}
Han, S. and Bauchau, O.~A., \enquote{On the global interpolation of motion,}
  \emph{Computer Methods in Applied Mechanics and Engineering}, 337:352--386,
  2018.

\bibitem[{Hughes(1986)}]{hughes86}
Hughes, P.~C., \emph{Spacecraft attitude dynamics}, John Wiley and Sons, New
  York, 1986.

\bibitem[{Junkins et~al.(2011)Junkins, Majji, Macomber, Davis, Doebbler, and
  Nosterk}]{junkins11}
Junkins, J.~L., Majji, M., Macomber, B., Davis, J., Doebbler, J., and Nosterk,
  R., \enquote{Small Body Proximity Sensing with a Novel HD 3D Ladar System,}
  \emph{Advances in the Astronautical Sciences}, 141:341--354, 2011.

\bibitem[{Maimone et~al.(2007)Maimone, Cheng, and Matthies}]{maimone07}
Maimone, M., Cheng, Y., and Matthies, L., \enquote{Two years of visual odometry
  on the mars exploration rovers,} \emph{Journal of Field Robotics},
  24(3):169--186, 2007.

\bibitem[{Majji et~al.(2011)Majji, Davis, Doebbler, Junkins, Macomber, Vavrina,
  and Vian}]{majji11}
Majji, M., Davis, J., Doebbler, J., Junkins, J., Macomber, B., Vavrina, M., and
  Vian, J., \enquote{Terrain mapping and landing operations using vision based
  navigation systems,} in \emph{AIAA Guidance, Navigation, and Control
  Conference}, page 6581, 2011.

\bibitem[{Matthies and Shafer(1987)}]{matthies87}
Matthies, L. and Shafer, S., \enquote{Error modeling in stereo navigation,}
  \emph{IEEE Journal on Robotics and Automation}, 3(3):239--248, 1987.

\bibitem[{Mortari et~al.(2007)Mortari, Markley, and Singla}]{mortari07}
Mortari, D., Markley, F.~L., and Singla, P., \enquote{Optimal linear attitude
  estimator,} \emph{Journal of Guidance, Control, and Dynamics},
  30(6):1619--1627, 2007.

\bibitem[{Mozzi(1763)}]{mozzi1763}
Mozzi, G., \emph{Discorso matematico sopra il rotamento momentaneo dei corpi},
  Donate Campo, 1763.

\bibitem[{Murray et~al.(1994)Murray, Li, and Sastry}]{murray94}
Murray, R.~M., Li, Z., and Sastry, S., \emph{A Mathematical Introduction to
  Robotic Manipulation}, CRC Press, 1994.

\bibitem[{Qian et~al.(2020)Qian, Charland-Arcand, and Forbes}]{qian20}
Qian, D., Charland-Arcand, G., and Forbes, J.~R., \enquote{TWOLATE: Total
  Registration of Point-Clouds Using a Weighted Optimal Linear Attitude and
  Translation Estimator,} in \emph{2020 IEEE Conference on Control Technology
  and Applications (CCTA)}, pages 43--48, IEEE, 2020.

\bibitem[{Selig(2007)}]{selig07}
Selig, J.~M., \enquote{Cayley maps for {SE(3)},} in \emph{12th International
  Federation for the Promotion of Mechanism and Machine Science World
  Congress}, page~6, London South Bank University, 2007.

\bibitem[{Stuelpnagel(1964)}]{stuelpnagel1964}
Stuelpnagel, J., \enquote{On the parametrization of the three-dimensional
  rotation group,} \emph{SIAM review}, 6(4):422--430, 1964.

\bibitem[{Wong and Majji(2016)}]{wong16}
Wong, X.~I. and Majji, M., \enquote{A structured light system for relative
  navigation applications,} \emph{IEEE Sensors Journal}, 16(17):6662--6679,
  2016.

\bibitem[{Wong et~al.(2018)Wong, Singla, Lee, and Majji}]{wong18}
Wong, X.~I., Singla, P., Lee, T., and Majji, M., \enquote{Optimal Linear
  Attitude Estimator for Alignment of Point Clouds,} in \emph{2018 IEEE/CVF
  Conference on Computer Vision and Pattern Recognition Workshops (CVPRW)},
  pages 1577--15778, IEEE, 2018.

\bibitem[{Yang et~al.(2020)Yang, Shi, and Carlone}]{yang20}
Yang, H., Shi, J., and Carlone, L., \enquote{Teaser: Fast and certifiable point
  cloud registration,} \emph{IEEE Transactions on Robotics}, 37(2):314--333,
  2020.

\end{thebibliography}

\end{document}